\documentclass[10pt,letterpaper]{article}
\usepackage[top=0.85in,left=2.75in,footskip=0.75in]{geometry}

\usepackage{amsmath,amssymb}

\usepackage{changepage}

\usepackage{textcomp,marvosym}

\usepackage{cite}

\usepackage{nameref,hyperref}

\usepackage[right]{lineno}

\usepackage[nopatch=eqnum]{microtype}

\usepackage[table]{xcolor}

\usepackage{array}

\newcolumntype{+}{!{\vrule width 2pt}}

\newlength\savedwidth

\newcommand\thickhline{\noalign{\global\savedwidth\arrayrulewidth\global\arrayrulewidth 2pt}%
\hline
\noalign{\global\arrayrulewidth\savedwidth}}

\raggedright
\usepackage[aboveskip=1pt,labelfont=bf,labelsep=period,justification=raggedright,singlelinecheck=off]{caption}

\makeatletter
\renewcommand{\@biblabel}[1]{\quad#1.}
\makeatother

\usepackage{lastpage,fancyhdr,graphicx}
\usepackage{epstopdf}
\fancyheadoffset[L]{2.25in}
\fancyfootoffset[L]{2.25in}
\usepackage{expex}
\lingset{everygla=, belowglpreambleskip=0ex, aboveglftskip=-0.4ex}

\usepackage{comment}

\usepackage{float}

\usepackage{polyglossia}
\usepackage{fontspec}
\setdefaultlanguage{english} 
\setotherlanguage{urdu} 
\newfontfamily\urdufont{Amiri-Regular.ttf}[Script=Arabic, Language=Urdu]

\begin{document}
\vspace*{0.2in}

\begin{flushleft}
{\Large
\textbf\newline{Multi-Task Learning by using Contextualized Word Representations for Syntactic Parsing of a {\color{black}Morphologically Rich} Language} 
}
\newline
\\
Toqeer Ehsan\textsuperscript{1*},
Miriam Butt\textsuperscript{2},
Sarmad Hussain\textsuperscript{3},
Hassan Alhuzali\textsuperscript{4},
Ali Al-Laith\textsuperscript{5}
\\
\bigskip
\textsuperscript{1}Reliable Intelligence Team, VTT Technical Research Centre of Finland Ltd., 02150 Espoo, Finland.
\\
\textsuperscript{2}Department of Linguistics, University of Konstanz, 78464 Konstanz, Germany.
\\
\textsuperscript{3}Center for Language Engineering, Al-Khawarizmi Institute of Computer Science, University of Engineering and Technology, 54890 Lahore, Pakistan.
\\
\textsuperscript{4}Department of Computer Science and Artificial Intelligence, Umm Al-Qura University, 24382 Makkah, Saudi Arabia.
\\
\textsuperscript{5}Computer Science Department, Copenhagen University, 2300 Copenhagen, Denmark.
\smallskip

* toqeer.ehsan@vtt.fi
%
%
\end{flushleft}
\section*{Abstract}

{\color{black}
We address the challenge of syntactic parsing for Urdu, a morphologically rich language, and present state-of-the-art results for both constituency and dependency parsing. This paper offers four major contributions: 1) the conversion of the CLE-UTB phrase structure treebank into a dependency treebank by developing language-specific head-word and phrase-to-dependency label mapping rules; 2) a novel sequence labeling scheme that transforms the parsing task into a unified representation; 3) the training of contextualized word representations on a large 220 million tokens Urdu corpus collected from the web; and 4) development of parsing framework using two learning paradigms, single-task and multi-task learning. Several post-processing rules are applied to improve the quality of the automatically converted dependency structure treebank. The proposed sequence labeling scheme enables the use of a shared architecture that learns the syntactic structures from both grammatical structures simultaneously and hence improves generalization. Experiments show that the multi-task learning setup significantly enhances parsing performance, achieving an F1 score of 91.39 for constituency parsing (an improvement of 3.29 points) and a labeled attachment score of 85.69 for dependency parsing (an improvement of 1.49 points). These results demonstrate that learning cross-task representations provides measurable benefits and advances the state of syntactic parsing for Urdu.
}



\section{Introduction}
\label{sec-1}

Urdu is a {\color{black}morphologically rich} language, written in a version of the Arabic script from right to left. As an Indo-Aryan language, it is mainly spoken in the South Asian region. There are over 160 million speakers worldwide who use Urdu as their first and second language \cite{simons2017ethnologue}. It holds several significant linguistic properties that include rich morphology \cite{hussain2004finite}, a complex case system \cite{butt2004status, ahmed2009spatial}, complex predicate structure \cite{butt2001complex, butt1995structure} and flexible word-order \cite{raza2011argument}. The distinctive characteristics of Urdu pose challenges for natural language processing. Despite being considered a low resource language, a number of efforts have been undertaken in the past years to develop essential linguistic resources, enabling the performance of various NLP tasks. These include the creation of Urdu corpora, morphological analysis \cite{hussain2004finite, bogel2008developing}, tokenization and word segmentation \cite{durrani2010urdu, zia2018urdu}, part of speech (POS) tagging \cite{ahmed2015cle, khan2019urdu}, sequence chunking \cite{ehsan2021improving} and syntactic parsing \cite{bhat2017improving, ehsan2019analysis, ehsan2021development, ehsan2020dependency}. Urdu remains under-resourced despite these efforts emphasizing the need for continued development of linguistic resources and tools to achieve computational results that can compete with more linguistically resourced languages.

A phrase=structure (PS) treebank illustrates the constituency structure of clauses and their hierarchical organization in a sentence. The information of arguments is encoded at phrase level in the form of linear order. A {\color{black}PS} parser provides the information of clauses in sentences, but it lacks grammatical relations. However, functional labels can be attached to phrase labels to represent grammatical roles. On the other hand, the dependency=structure (DS) focuses on functional dependencies between lexical items in clauses and dependency relations between constituents. It highlights grammatical relations between predicates and their arguments which are essential to extract the information about event participants in the applications of natural language understanding (NLU). Urdu has both types of treebanks available that vary in size and annotation schemes. A promising approach is to convert existing {\color{black}PS} treebanks into common {\color{black}DS}. This conversion aims to facilitate the training of high-quality parsers that can capture both constituency and dependency information. Treebanks are essential resources for a range of NLP tasks. {\color{black} Several treebanks have been developed using the PS \cite{marcus1993building, maamouri2004penn, xue2005penn, nguyen2015vietnamese, nguyen2017ensuring} and the DS \cite{liu2006building, hajivc2017prague, brants2002tiger, silveira2014gold}.} The development of the Penn Treebank gave new perspectives and inspirations for the creation of syntactic resources for other languages and corpus-based analysis. 

Multi-task learning is an approach that involves training a single model to handle multiple related tasks simultaneously. {\color{black}It leverages common knowledge and cross-representations to improve overall performance}. In recent years, multi-task learning has found applications in various natural language processing tasks \cite{zhang2022survey} and demonstrated its applicability in sentiment analysis \cite{tan2023sentiment, zhao2023aspect}, named entity recognition \cite{huang2022multitask, zhang2022biomedical}, semantic role labeling \cite{arora2022multi}, {\color{black}extraction of biomedical relations} \cite{moscato2023multi}, polarity classification \cite{zhao2023multi}, sarcasm and toxic comments detection \cite{tan2023sentiment, nelatoori2023multi}, speech recognition \cite{wan2023grammar} and syntactic parsing \cite{strzyz2019sequence, fernandez2022multitask}.
In this work, we present the conversion of an Urdu {\color{black}PS} treebank \cite{ehsan2021development} into a {\color{black}DS}, utilizing the Universal Dependencies (UD V2) label sets. To achieve an equivalent {\color{black}DS}, we devised a head-word model and created a mapping from phrase labels to dependency labels. {\color{black}In addition}, various post-conversion rules have been employed to ensure accurate conversion. In natural language parsing, constituency and dependency parsing are considered disjoint tasks and executed independently \cite{kiperwasser2016simple, dozat2016deep, ma2018stack, kitaev2018constituency}. However, there is potential {\color{black}to improve} both constituency and dependency parsing by incorporating cross-structure representations into the learning process \cite{strzyz2019sequence, fernandez2022multitask}. 

For the constituency structure, we enhanced {\color{black}CLE-UTB labeling scheme}, resulting in improved scores. Concerning the {\color{black}DS}, we employed the sequence labeling approach from \cite{strzyz2019viable}, initially introduced in \cite{spoustova2010dependency}. {\color{black}Prior to multi-task learning, we performed single-task learning-based parsing experiment to achieve the results for constituency and dependency parsing independently. We developed a neural parsing framework that is based on Long Short-Term Memory (LSTM) networks. Bidirectional LSTM (BLSTM) networks have proven to be highly effective in solving sequence labeling problems}. The multi-task learning paradigm transforms the constituency and dependency parsing into a sequence labeling problem aligning well with the capabilities of BLSTM networks for neural modeling. Various features have been incorporated into the experiments including character embeddings and convolutions, POS tags, and pre-trained word representations. {\color{black}To perform transfer learning, we trained two types of word representations, context-free embeddings \cite{mikolov2013efficient} and deep contextualized word representations (ELMo) \cite{peters2018deep}. The single-task constituency parser achieved the highest F-score of 90.94 when trained with three hidden LSTM layers along with POS tags and embeddings from the third layer of ELMo representations}. Similarly, the single-task dependency parser obtained a labeled attachment score (LAS) of 85.36 incorporating POS tags, Word2Vec and ELMo embeddings. {\color{black}The multitask learning-based parser achieved an F-score of 91.39 with an improvement of 3.29 points compared to the state-of-the-art score for the CLE-UTB and with an improvement of 3.53 points for function labeling accuracy. The multi-task dependency parser improved the parsing scores by 1.49 points and achieved a labeled attachment score of 85.69}. The single-task constituency and dependency scores presented in this work are state-of-the-art results for the CLE-UTB which are further improved by learning cross-structure representations by employing multi-task learning.

The remainder of the paper is organized as follows. Section~\ref{sec-2} briefly describes the compatibility of the {\color{black}PS} CLE-UTB with the {\color{black}DS} and related work. Section~\ref{sec-3} presents the conversion approach by providing the details of head-word model, phrase-to-dependency label mappings, and post-conversion rules. Section~\ref{sec-4} describes the sequence labeling schemes for both PS and DS treebanks. Section~\ref{sec-5} presents neural parsing models that include single-task and multi-task learning models along with the transfer learning approach. Section~\ref{sec-6} presents parsing results and their interpretation, and the conclusion of the work is presented in Section~\ref{sec-7}. 

\section{Related Work} 
\label{sec-2}

{\color{black}Currently, there are two prominent PS treebanks available for Urdu, Urdu.Kon-TB \cite{abbas2012building} and CLE-UTB \cite{ehsan2021development}. Both treebanks are quite different with respect to their structure, annotation scheme, annotation depth, genre coverage, and size. This section provides analysis of their structure and suitability for multi-task learning parsing}.

\begin{figure}[ht] \centering
\scalebox{0.40}{
\includegraphics[scale=0.36]{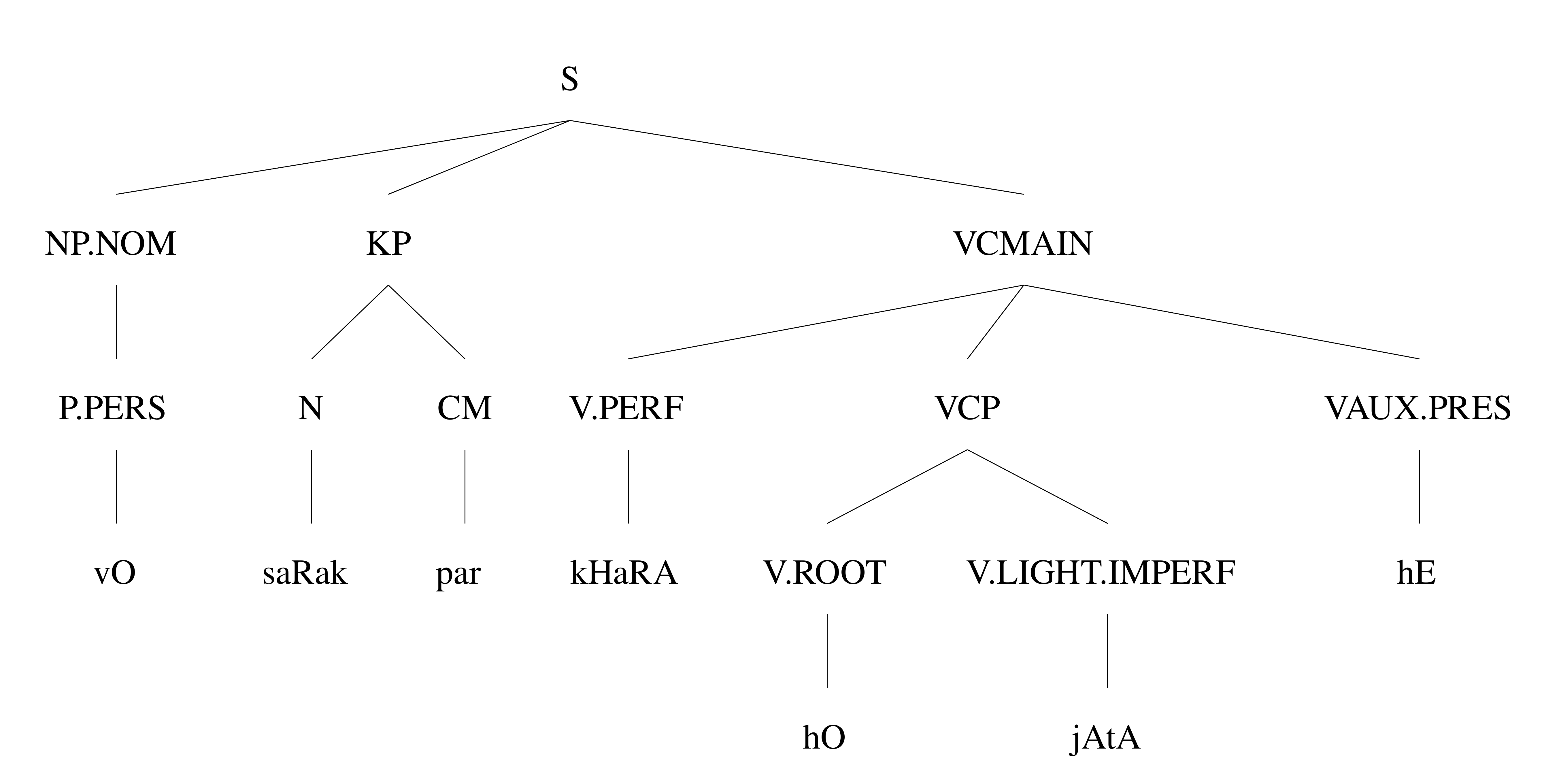}
 }	\\ 
\scalebox{1.0}{	
\setlength\tabcolsep{1.8pt}
\begin{tabular}{llllll}
\\vO&saRak=par&kHaRa&hO&jAtA&hE\\
Pron.3.M.Sg&road.F.Sg=Loc& stand.M.Sg& be.Imper.Sg& go.Pres.M.Sg& be.Pres.3.Sg\\
\multicolumn{6}{l}{`He use to stand on the road'.} 
\end{tabular}
}
\caption{A sample {\color{black}PS} parse tree from the Urdu.KON-TB.}
\label{fig-0.1}
\end{figure}

The Urdu.KON-TB has multiple annotation layers, including POS tags, phrase labels, and functional tags. {\color{black}The phrase labels have morphological information attached with them. The phrase label concatenates the information of tense, case, aspect, and modality. The treebank has been annotated using a large annotation scheme containing 26 phrase labels. However, the treebank contains only 1,400 sentences with an average of 13.73 tokens per sentence. The corpus has been collected mainly from the Urdu Wikipedia (https://ur.wikipedia.org) and the Jang newspaper (https://jang.com.pk)}. Figure~\ref{fig-0.1} demonstrates a sample annotated sentence from the Urdu.KON-TB. The Urdu text in all examples has been written using a transliteration scheme proposed in \cite{kamran2010transliterating}.

The parse tree in Figure~\ref{fig-0.1} shows the attachment of three clauses under the `S' phrase label. The `S' label depicts the annotation of the whole sentence that has an NP.NOM, a KP, and a VCMAIN. The NP.NOM defines the nominal case along with a noun phrase. The POS tag for personal pronoun is P.PERS which marks the pronoun by depicting its type as well. The nominal case represents the nominal subject. The second clause annotates a case marking clause with the KP label. The annotation of the case phrase is quite flat as noun and case marker appear at the same level. The third clause defines the verbal structure with the VCMAIN label. {\color{black}It contains a main verb annotated with V.PERF that shows the verb and its tense, a VCP clause that contains the root of the verb and a light verb, and finally an auxiliary verb by using a VAUX label along with the tense information}. The verbal construction is quite complicated as it demonstrates a main verb \textit{kHaRA} `stand.M.Sg' with a root \textit{hO} `be.Imper.Sg' and a light verb \textit{jAtA} `go.Pres.M.Sg'. The verbal construction actually makes a complex predication to denote adjective+verb construction. 
{\color{black}This annotation lacks annotation of different order of arguments within a complex predicate. Nouns or adjectives do not always appear with verbs; therefore, a different annotation approach is required that has the ability to accommodate the flexible word-order of the language}. The Urdu.KON-TB divides the verbal structure into four categories that include a VCMAIN; which annotates the main clause, a VCP; to annotate complex predicates, a VIP; that marks infinitive verb phrase, and a VP for simple verb phrases. The Urdu.KON-TB has a large annotation label set that contains the duplicate labels for questions like NPQ, KPQ, QWP, ADJPQ, ADVPQ, SBARQ and SQ.

{\color{black}The linguistically rich annotation scheme of the Urdu.KON-TB provides a deeper linguistic insight into the language. However, it is a small treebank that has been annotated using a complex labeling scheme}. Although statistical and neural parsing models require large amounts of data for better performance, Urdu.KON-TB has been analyzed to perform dependency parsing by converting it to dependency structure in \cite{munir2017dependency}. The well-known MaltParser \cite{nivre2007maltparser} was used to perform dependency parsing. Despite the richer annotation scheme, the conversion to dependency structure was incorrect due to the deeper hierarchical structure of the parse trees. Based on experiments, the authors concluded that the Urdu.KON-TB is not suitable for dependency parsing.

\begin{figure}[ht] \centering
\scalebox{0.45}{
\includegraphics[scale=0.45]{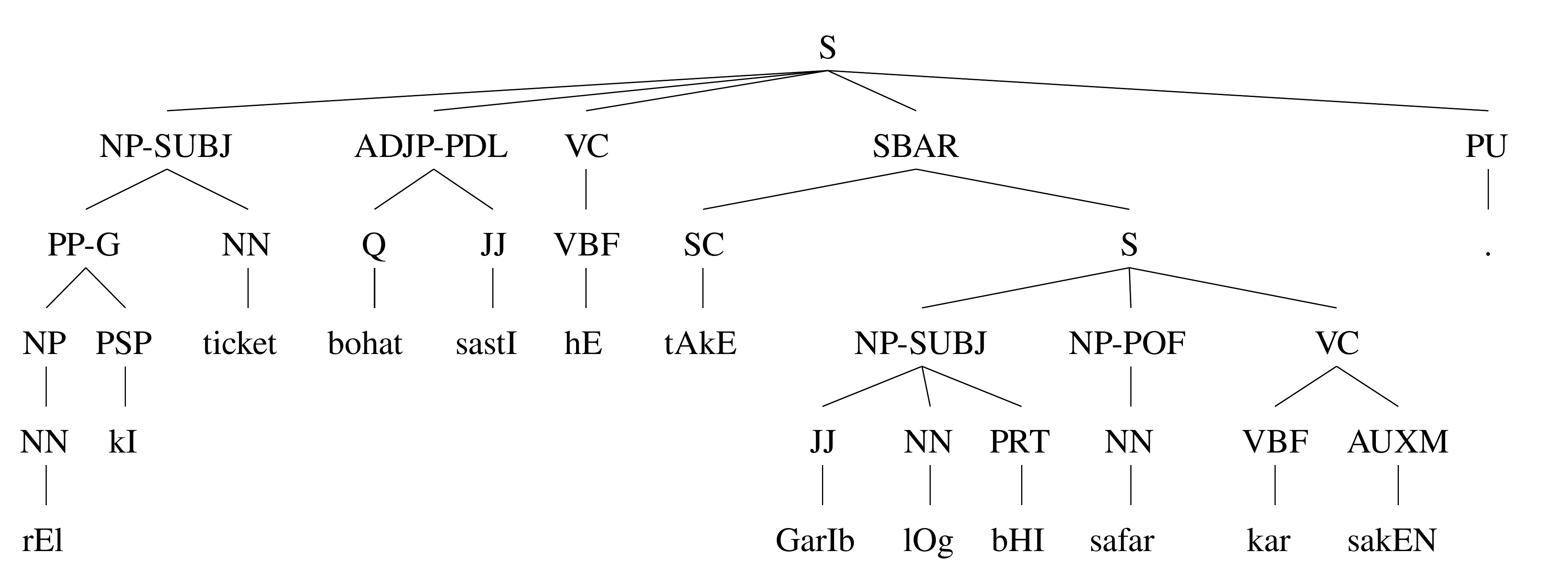}
 }	\\ 
\scalebox{1.0}{	
\setlength\tabcolsep{1.5pt}
\begin{tabular}{lllllll}
\\rEl=kI&ticket&bohat&sastI&hE&tAkE&GarIb\\
train.F.Sg.Gen&ticket.F.Sg.Nom&very&cheap&be.Pres.3.Sg&SC&poor.Nom \\
lOg&bHI&safar &kar & sakEN &. &\\
people.Pl.Nom&also.PRT&travel.Nom & do & can.Pres.Pl.Mod &Punct&\\
\multicolumn{7}{l}{`The train ticket is very cheap so that poor people can also travel'.} 
\end{tabular}
}
\caption{A sample {\color{black}PS} parse tree from the CLE-UTB.}
\label{fig-1}
\end{figure}

On the other hand, the CLE-UTB has a relatively flat annotation structure. {\color{black}The flat structure is helpful for annotating the flexible word order of the language by attaching arguments in any order in a parse tree that is a key attribute of the dependency structure}. The annotation scheme of the CLE-UTB has the ability to represent frequently observed syntactic constructions of the language, including flexible word-order, complex predicate structure, question phrases, subordination, conjunctions, genitives, and relative clauses. Additionally, the CLE-UTB has a set of functional labels to mark grammatical relations. Figure~\ref{fig-1} presents a sample annotated sentence from the CLE-UTB.

The PS parse tree in Figure~\ref{fig-1} shows the annotation of a genitive case and a copula construction in the independent clause. The genitive case is annotated by using a post-positional phrase (PP) and a function label `G'. The PP-G phrase is a component of the nominal subject that has \textit{ticket} as the head word. The independent clause further annotates a copula construction that marks an adjective phrase (ADJP) with a functional label PDL (Predicate Link). The PDL construction makes a link with the predicate, which, in this case, is a copula verb \textit{hE} `be.Pres.3.Sg'.

The subordinate clause, on the other hand, contains a nominal subject and a complex verbal structure. The NP-SUBJ phrase contains an intensifier \textit{bHI} `also' in it. The nominal subject is followed by a complex predicate structure that has been marked by using a {\color{black}Part-of-Function (POF)} label. The noun phrase along with POF provides a verbal meaning generally followed by a Verbal Complex (VC). In the complex predicate structure, the predicate contains a light verb, \textit{kar} `do' in this case, which is further followed by a modal auxiliary verb \textit{sakEN} `can.Pres.Pl.Mod'.

{\color{black}It is important to note that all the arguments and predicates are attached at the clause level in the parse tree shown in Figure~\ref{fig-1}. This annotation scheme of the CLE-UTB allows the annotation of sentences having different word orders. 
The functional layer makes the annotation flexible, and predicates can be marked irrespective of their order in the sentence. Similarly, copula constructions can be annotated independently of their position. Predicates and their arguments including subjects, objects, obliques, conjuncts, infinitive clauses, subordinate and relative clauses are annotated without restricting them to have specific positions in a parse tree}. 

The annotation scheme of the CLE-UTB is flat and has fewer labels compared to the Urdu.KON-TB. The CLE-UTB's labels including POS tags, phrase and functional labels are also compatible with existing resources like the well-known Penn Treebank. However, some POS tags and phrase labels have been added to represent the grammatical structure of Urdu. The flat annotation approach makes it compatible with the dependency structure \cite{ehsan2020dependency,ehsan2021development}. The CLE-UTB is sufficiently large that has 7,854 sentences having text from fifteen different genres. In this work, we have transformed the CLE-UTB from phrase structure to dependency structure and performed multi-task learning-based parsing.

{\color{black}
In recent years, Urdu NLP has witnessed significant advancements in a variety of core tasks. In semantic parsing, the first semantically annotated meaning bank for Urdu was created to enable neural parsing and generation, leveraging cross-lingual alignment and extensive data augmentation yielded a major boost in parsing accuracy \cite{amin2025semantic}. In machine translation, systematic benchmarks show that specialized multilingual models can outperform general-purpose LLMs for English-Urdu translation. For example, the IndicTrans2 model achieved the highest translation quality on multiple test sets, significantly outperforming both a GPT-3.5-based translation model and a conventional bilingual Transformer \cite{basit2024challenges}. A recent study shows that it is crucial to use optimized part-of-speech (POS) tagging to improve the performance of Urdu text classification. A new POS tagging technique, coupled with SVM variants and ensemble models, led to significant improvements in classification accuracy, achieving over 98\% average scores on several evaluation measures on a four-domain corpus \cite{ishaq2025development}. Another recent research explored the use of Large Language Models (LLMs) to address dynamic label problems in large-scale multi-label text classification (LMTC). A recent strategy focused on DyLas (Dynamic Label Alignment Strategy) applies counterfactual analysis to align labels without the requirement of retraining a model and demonstrated performance improvements with different datasets and open and closed source LLMs \cite{ren2025dylas}.

Low-resource challenges are also being addressed by new resources for higher-level tasks such as summarization and question answering. The first multimodal Urdu summarization dataset (UrduMASD) was introduced in 2024 to enable abstractive summaries from videos and text \cite{faheem2024urdumasd}. Similarly, a large-scale QA corpus (UQA) was constructed by translating the SQuAD2.0 benchmark \cite{rajpurkar2018know} into Urdu using a span-preserving method \cite{arif2024uqa}, and state-of-the-art multilingual transformer models have already achieved strong performance on this dataset. Conversational generation has made strides recently in overcoming the basic issues of semantic ambiguity and contextual coherence through better transformer-based models. A novel approach combining pre-trained word embeddings with character information and sparsity-regularized Softmax loss resulted in substantial improvement in BLEU and distinct scores with better coherent and contextually relevant responses on the STC dataset \cite{lv2024enhancing}. 

Previous work has extended clickbait detection research to the low-resource Urdu language, which was not addressed in this task previously. Using sentence embeddings with deep learning methods, researchers achieved 88\% accuracy in identifying clickbait headlines, much higher than typical machine learning baselines \cite{muqadas2025deep}. The advent of large language models is beginning to benefit Urdu. Researchers have introduced UrduLLaMA 1.0, an 8-billion-parameter model tailored to Urdu by continual pretraining and LoRA fine-tuning on Urdu data, which has set new benchmarks on several Urdu NLP tasks by substantially improving prior state-of-the-art results \cite{fiaz2025urdullama}. 

In the field of speech processing, recent multilingual approaches have greatly improved Urdu speech recognition and synthesis. On the recognition side, state-of-the-art models (e.g. OpenAI Whisper, Meta's MMS and SeamlessM4T) have been fine-tuned and benchmarked on Urdu, aided by the release of the first large-scale Urdu conversational speech dataset \cite{arif2024we}. These evaluations show that fine-tuned models can achieve relatively low word error rates for Urdu (whisper performing best in conversational speech), although challenges such as accent variability and the need for robust text normalization persist \cite{arif2024we}. 

In text-to-speech, the lack of Urdu training data is being addressed by transfer learning: for instance, a study constructed an Urdu speech corpus from audio books and then used a pre-trained neural Text-to-Speech (TTS) model (trained on a high-resource language) to generate intelligible Urdu speech \cite{jamal2022exploring}. The resulting system demonstrated satisfactory naturalness despite the limited data, although there is still room for improvement. Meanwhile, Urdu conversational agents have begun to emerge in real-world applications. A notable example is the ``SAATHI" virtual assistant, which provides an intuitive Urdu voice interface for elderly users, helping with daily tasks such as reminders of medications, information access, and even offering empathetic conversations for companionship \cite{kumar2022saathi}. Recent contributions to Urdu speech processing have introduced a TTS system powered by Deep Learning techniques using Tacotron 2 with WaveGlow to generate natural-sounding speech from Urdu text. The system achieved a Mean Opinion Score (MOS) of 3.76, overcoming previous research on Urdu TTS that lacked advanced neural architectures \cite{saba2022urdu}. Such developments illustrate how advances in language resources and models are making Urdu NLP technologies more robust and accessible across both text and speech domains.

}

\section{PS to DS Conversion}
\label{sec-3}
{\color{black}
We conducted the conversion process in three steps. In the first step, we generated an intermediate dependency representation by implementing a head-word model. This representation contains dependency arcs along with phrase and POS tags. In the second step, we performed the phrase-to-dependency label mapping to transform the intermediate representation into the dependency structure. Finally, in the third step, we applied post-conversion rules to generate the final dependency treebank. This section outlines the phrase-to-dependency structure conversion by detailing the Urdu head-word model, phrase-to-dependency label mapping, post-conversion rules, and the conversion evaluation.}

\subsection{Urdu Head-word model}
\label{sec-3-1}
In {\color{black}DS}, a head-word is marked with the core dependency labels. A head-word is a dominant lexical item in a phrase. A head-word algorithm finds head-words from clauses by implementing language-dependent rules \cite{magerman1995statistical}. We have proposed a head-word model for the CLE-UTB treebank. In our treebank conversion process, we refined an existing algorithm (https://github.com/Luolc/CTB2Dep) \cite{techreport}. Table~\ref{tab-1} shows label-wise details of the proposed head-word model.

\begin{table}[h]
\centering
\small
\caption{Head-word model for the CLE-UTB treebank.}
\label{tab-1}
\begin{center}
\begin{tabular}{@{}lll@{}}
\hline
Phrase labels & Direction & Priority \\ 
\thickhline
VC & left & VBF, VBI, AUXA, AUXM, AUXP, AUXT, VC, NEG\\
PP & left & NP, S, QP, NNP, NN, PP, PSP \\
NP & right & NP, NNP, NN, PRP, PRR, S \\
ADJP & right & ADJP, JJ, Q, QP, RB \\
QP & right & QP, Q, CD, OD, FR, QM, JJ \\
ADVP & right & ADVP, RB, NP, NN \\
PREP & right & NP, NNP, NNP, PREP \\
DMP & right & PDM, PRP, PRT \\
FFP & left & FF, NNP, NN \\
S & left & VC, S, SBAR, NP, ADJP, QP, NNP, NN, PRP\\
SBAR & left & S, SBAR, SCK\\
\hline
\end{tabular}
\end{center}
\end{table}

{\color{black}Table~\ref{tab-1} presents the word finding direction and the priority list for each phrase label}. A verb complex generally contains a main verb followed by one or more auxiliary verbs. Therefore, the head-finding direction is from left to right with respect to the priority of labels. A VC mostly has finite verbs as main verbs; therefore, the POS tags VBF and VBI have high priorities compared to the auxiliaries. The verbal constituent in (\ref{g-1}) demonstrates a typical Urdu verbal structure which contains the main verb \textit{likH} `write', a light verb \textit{liyA} `take.Perf.M.Sg' and a tense auxiliary verb \textit{hE} `be.Pres.3.Sg'. The main verb should be the head of the phrase and have the core dependency label that appears on the left-hand side. {\color{black}It is important to note that Urdu is written from right-to-left, but its internal representation is a sequence of left-to-right Unicodes, therefore, the directions in our head-word model have been devised for internal representations of the writing script.}

\ex
\begingl 
\label{g-1}
\gla likH liyA hE//
\glb write.Imperf.Sg take.Perf.M.Sg be.Pres.3.Sg//
\glft ~`has written'//
\endgl
\xe

Similarly, post-positional phrases (PPs) have left heads as they contain at least one noun phrase that appears before a case marker. Therefore, the PP has NP to have the highest priority followed by other clause labels. In (\ref{g-2}), a PP sequence contains a noun phrase \textit{surx mEz} `red table' followed by a locative case marker \textit{par} `on'. {\color{black}The head-word of the noun phrase becomes the head of the PP phrase. The general direction of the head of the PP is from left-to-right. Besides NPs, a PP may have other phrases that have been added to its priority list.}

\ex
\begingl 
\label{g-2}
\gla surx mEz=par//
\glb red.Sg table.M.Sg=Loc//
\glft ~`on the red table'//
\endgl
\xe

The noun phrase shown in (\ref{g-3}) contains a genitive case, and hence an internal post-positional phrase. The head of the NP is the word \textit{pensil} `pencil' which appears on the right-hand side of the phrase. Similarly, the NP sequence presented in (\ref{g-4}) shows a canonical NP construction that contains the noun \textit{Am} `mango' as head-word after a noun modifier \textit{mItHA} `sweet'. The head-finding direction of noun phrases is from right-to-left. A noun phrase can also have internal NPs, common and proper nouns and pronouns that all are right headed.

\ex
\begingl 
\label{g-3}
\gla sAim=kI pensil//
\glb Saim.M.Sg=Gen pencil.F.Sg.Nom//
\glft ~`Saim's pencil'//
\endgl
\xe

\ex
\begingl 
\label{g-4}
\gla mItHA Am//
\glb sweet.Sg mango.M.Sg.Nom//
\glft ~`sweet mango'//
\endgl
\xe

The noun modifiers also have right heads as shown in (\ref{g-5}) and (\ref{g-6}). An adjective phrase, in the annotation, can have quantifiers, cardinals, ordinals, and main adjectives. The order of an adjective phrase is similar to that of canonical NPs. The head modifier appears on the right-hand side of the constituent. In the same way, the quantifier phrase also has right heads, as shown in (\ref{g-6}). 

\ex
\begingl 
\label{g-5}
\gla buhat mazbUt//
\glb very strong.M.Sg.Adj//
\glft ~`very strong'//
\endgl
\xe

\ex
\begingl 
\label{g-6}
\gla buhat kam//
\glb very less.M.Sg.Adj//
\glft ~`very less'//
\endgl
\xe

An adverbial phrase (ADVP) mainly contains a single lexical entry for an adverb as presented in (\ref{g-7}). In some cases, the ADVP contains an infinitive clause, but the head-word remains on the right-hand side of the constituent. 

\ex
\begingl 
\label{g-7}
\gla acAnak//
\glb suddenly.Adv//
\glft ~`suddenly'//
\endgl
\xe

Urdu also has prepositional phrases, but they are not frequent in corpora. A prepositional phrase contains an NP that follows the preposition clitic, which makes it right headed as shown in (\ref{g-8}). The preposition clitic \textit{fI} `per' appears before a noun phrase making the noun the head of the constituent.

\ex
\begingl 
\label{g-8}
\gla fI gHanTA//
\glb per hour.M.Sg.Nom//
\glft ~`per hour'//
\endgl
\xe

\begin{figure}[ht] \centering
\scalebox{0.45}{
\includegraphics[scale=0.45,clip, trim=0.9cm 0cm 5cm 6cm]{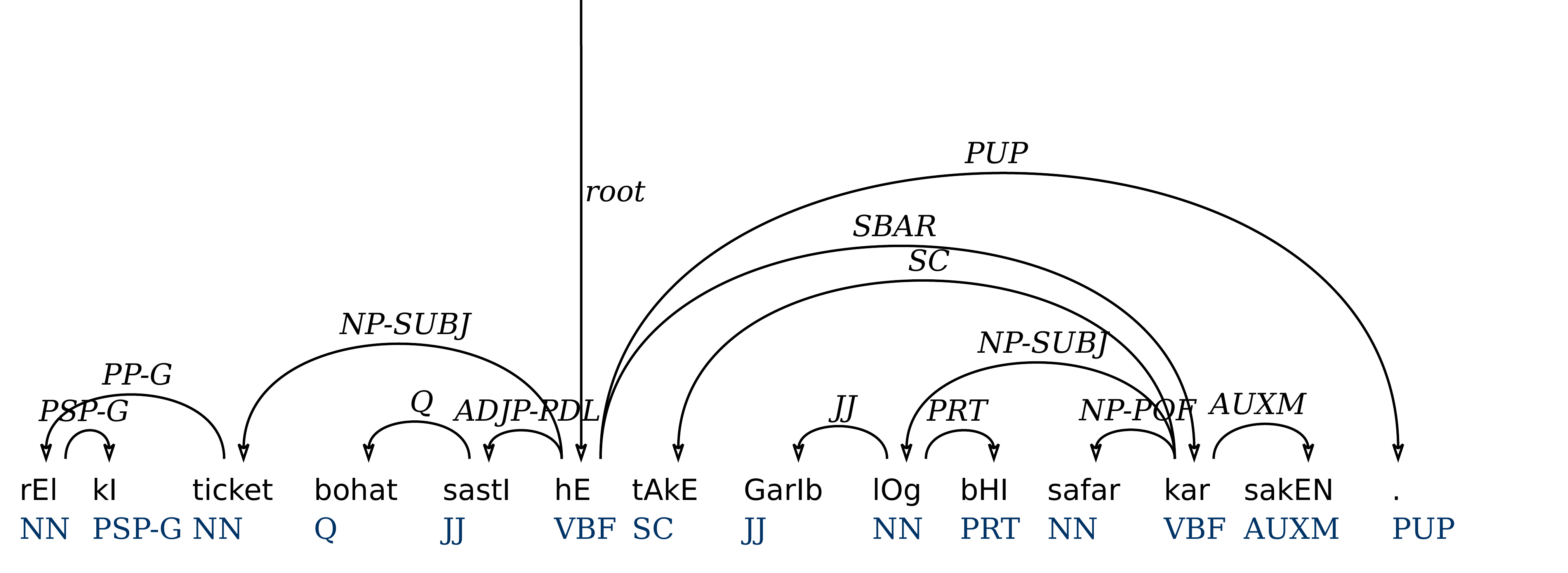}
}
\caption{An intermediate dependency representation from PS parse tree of Figure~\ref{fig-1} after head identification.}
\label{fig-2}
\end{figure}

A demonstrative phrase (DMP) contains a demonstrative pronoun followed by a particle, making its head-finding method straight-forward from right-to-left. Similarly, foreign fragment phrase (FFP) annotates the foreign language text segments, generally from English and Arabic in the treebank. {\color{black}All foreign words are assigned the POS tag of FF. We assume it to be a left-headed constituent. The `S' label is used to annotate clauses or a whole sentence. A sentence can have subordinate and coordinate clauses in it. The SBAR label is used to annotate subordinate clauses with the label `S'. The clause or sentence has the main verb as the head of the sentence. Therefore, it looks for a left head of the matrix clause as the head of the whole sentence. In Figure~\ref{fig-2}, a hierarchical analysis and parental annotation have been performed to determine heads for clause `S'. The parental annotation was helpful to achieve the context and phrase-to-dependency label mappings.}

Figure~\ref{fig-2} shows an intermediate dependency representation of the sentence in Figure~\ref{fig-1}. The representation is achieved by finding dependency arcs for head-words. The remaining items of the constituents show dependency arcs on the respective heads. This representation has phrase and functional labels as relations. Besides labels, the dependency arcs are correct with respect to the Universal Dependencies. The construction including subordinate clause, subjects, complex predicate, genitive case, and auxiliary verbs show dependency arcs on their respective head-words. 

\subsection{PS to DS Label Mappings}
\label{sec-3-2}
The intermediate dependency representation contains dependency arcs for head and non-head tokens. The non-head-words have POS tags as arc labels and heads have phrase labels as dependency relations. {\color{black}During the conversion process, we designed a refined version of the POS tag set that divides case markers and punctuation marks into multiple categories according to their syntax. The POS tag set originally had 35 tags, which are now extended to 44 tags. The Universal Dependencies (UD) V-2.0 label set has been used for dependency labels and POS tags. The universal POS tag set contains 17 tags (https://universaldependencies.org/u/pos). In addition to the UD guidelines, the Hindi-Urdu Treebank (HUTB) \cite{bhat2017hindi} has been analyzed for label mappings and post-conversion rules.}

{\color{black}In Figure~\ref{fig-2}, the noun \textit{ticket} is the head-word and shows a dependency relation with the NP-SUBJ label, which makes it intuitive that it is a subject. The noun \textit{rEl} `train' shows a dependency on the head with the PP-G label. The PP-G label is used to annotate post-positional phrases for genitive case markers (clitics). The Urdu script treats case marking clitics as independent tokens hence requires a dependency label.} The genitive case marker \textit{kI} has a dependency on the noun \textit{rEl} which is a possessor in the sentence. The quantifier \textit{bohat} `very' has a dependency on the adjective \textit{sastI} `cheap' with the `Q' label which is a POS tag. The adjective \textit{sastI} is the head-word with the core dependency label ADJP-PDL. Its appearance before a copula verb \textit{hE} makes a construction called a predicate link. We updated such copula constructions by implementing post-conversion rules.

The subordinate clause contains a subject and a complex predicate structure. The phrase labels NP-SUBJ and NP-POF, representing dependency relations for heads and POS tags, are labeled to show non-core relations. For example, the noun modifier \textit{GarIb} `poor' has a dependency on the noun \textit{lOg} `people' with the JJ label. The modal auxiliary is defined by the AUXM tag. The particle \textit{bHI} `also' is labeled with the PRT POS tag. The subordinate clause is marked with the SBAR label. {\color{black}It is important to note that phrase labels along with functional labels represent dependency relations of heads on the predicates, and POS tags depict relations of non-head items on constituent heads. Therefore, to generate dependency labels, the mapping to dependency labels should be performed for both phrase labels and POS tags.}

To achieve the context of constituents for accurate label mappings, we updated CLE-UTB to have parental annotations \cite{johnson1998pcfg}. A clause `S' may have different attachments that can be identified by using the parent label annotations along with its actual label. For example, if the clause `S' appears within a noun phrase, it is annotated as S\string^NP and if it is attached with a post-positional phrase, it would have the annotation as S\string^PP. The subordinate and coordinate clauses are also annotated with an `S' label that show S\string^SBAR for a subordinate clause and S\string^S for a coordinate clause. {\color{black}In addition to the parental annotation, tree levels are also annotated for VCs that are used to identify the roots of sentences. A sentence with many clauses has many verbal constructions in a hierarchical parse tree. The annotation of the level number attaches a number, for example, VC\string^1 represents the verbal construction at the first level, considering the sentence clause at the zeroth level.}

\subsubsection{Core Arguments}
\label{sec-3-2-1}
Table~\ref{tab-3} presents the mapping of phrase labels for core arguments to UD labels. The functional labels represent the core arguments for subject and object. The subordinate and coordinate clauses are marked with the {\tt ccomp} (clausal complement) label. Parental annotations have been used to identify clauses. The nominal obliques have been mapped on indirect objects {\tt iobj}. This mapping was further refined by employing the second rule described in Section~\ref{sec-3-3}. Infinitive clausal objects have been marked as indirect objects with the {\tt xcomp} label with respect to the HUTB. The labels with a `*' are achieved by additional rules described in Section~\ref{sec-3-3}.

\begin{table}[h]
\centering
\caption{The mapping of the CLE-UTB phrase labels on UD labels for core arguments.}
\label{tab-3}
\begin{center}
\setlength\tabcolsep{16pt}
\begin{tabular}{@{}lll@{}}
\hline
CLE-UTB lables & UD-2.0 labels & UD description \\ 
\thickhline
NP-SUBJ & nsubj & Nominal subject \\
PP-SUBJ & nsubj & Nominal subject \\
QP-SUBJ & nsubj & Nominal subject \\
NP-OBJ & obj & Object \\
PP-OBJ & obj & Object \\
QP-OBJ & obj & Object \\
S-OBJ & obj & Object \\
NP-OBL* & iobj & Indirect object \\
S-SUBJ & csubj & Clausal subject \\
S\textasciicircum{}S & ccomp & Clausal complement \\
SBAR & ccomp & Clausal complement \\
S\textasciicircum{}SBAR & ccomp & Clausal complement \\
S* & xcomp & Open clausal complement \\
\hline
\end{tabular}
\end{center}
* These dependency labels are achieved by post-conversion rules.
\end{table}

\subsubsection{Non-Core Dependents}
\label{sec-3-2-2}
Table~\ref{tab-4} presents the mapping of non-core dependencies. The non-core compulsory arguments are annotated by using the OBL functional label in the CLE-UTB that is mapped on the oblique dependency with the {\tt obl} tag. The functional label VOC has been mapped on {\tt vocative} label. Non-core finite clauses are marked using the {\tt advel} (adverbial clause modifier) label. Similarly, clauses with conjunctive participle are also marked with {\tt advel}. These types of clauses are attached under the SBAR label that are identified by using parental annotations. Adverbial phrases (ADVP) and nominal adjuncts have been mapped on adverbial modifier dependency relation using the {\tt advmod} label. The noun phrase, post-positional phrase, prepositional phrase, and quantifier phrase that do not have any functional label are considered adjuncts and are marked with an adverbial modifier. The POS tags RB and NEG for adverbs and negative adverbs, respectively, are also mapped on the {\tt advmod} label. 

\begin{table}[ht]
\centering
\caption{The mapping of the CLE-UTB phrase labels on UD labels for non-core dependents.}
\label{tab-4}
\begin{center}
\begin{tabular}{@{}lll@{}}
\hline
CLE-UTB lables & UD-2.0 labels & UD description \\
\thickhline
NP-OBL & obl & Oblique nominal \\
PP-OBL & obl & Oblique nominal \\
S-OBL & obl & Oblique nominal \\
ADVP-VOC & vocative & Vocative \\
NP-VOC & vocative & Vocative \\
S & advcl & Adverbial clause modifier \\
S\textasciicircum{}SBARSCK & advcl & Adverbial clause modifier \\
SBARSCK & advcl & Adverbial clause modifier \\
ADVP & advmod & Adverbial modifier \\
NP & advmod & Adverbial modifier \\
NP-ADJ & advmod & Adverbial modifier \\
PP & advmod & Adverbial modifier \\
PREP & advmod & Adverbial modifier \\
QP & advmod & Adverbial modifier \\
NEG & advmod & Adverbial modifier \\
RB & advmod & Adverbial modifier \\
ADVP-INJ & discourse & Discourse element \\
INJ & discourse & Discourse element \\
AUXA & aux & Auxiliary \\
AUXM & aux & Auxiliary \\
AUXP & aux & Auxiliary \\
AUXT & aux & Auxiliary \\
SCK & aux & Auxiliary \\
ADJP-PDL & cop & Copula \\
ADVP-PDL & cop & Copula \\
NP-PDL & cop & Copula \\
PP-PDL & cop & Copula \\
QP-PDL & cop & Copula \\
SC & mark & Marker \\
SCP & mark & Marker \\
\hline
\end{tabular}
\end{center}
\end{table}

Adverbial interjections are marked with the {\tt discourse} label. All types of auxiliary verbs and conjunctive participles are mapped on {\tt aux} label. The annotation of the CLE-UTB utilizes a PDL functional label to annotate predicate links with copula verbs; therefore, constructions with PDL labels are mapped on copula dependency with the {\tt cop} label. The labels for subordinate clitics are mapped on the {\tt mark} dependency label.

\subsubsection{Nominal Dependents}
\label{sec-3-2-3}
{\color{black}
Table~\ref{tab-5} presents the mapping of the nominal dependents. Genitive cases are marked as noun specifiers and possessors, and are annotated with the PP-G phrase in the PS treebank. For conversion, PP-G is marked as a nominal modifier using the {\tt nmod} label. Cardinals are marked as numeric modifier by using the {\tt nummod} label. The adjectival phrases and adjectival tags are mapped as adjectival modifiers. The adjectival clauses that are identified through parental annotations in the `S' clause, are mapped on the {\tt acl} label. The determiner label {\tt det} is used to represent demonstrative phrases, APNA particles, fractions, ordinals, quantifiers, and multiplicatives. Different types of pronouns, including demonstrative, relative demonstrative, reflexive, relative, and possessive, are also marked as determiners. The dependency label {\tt case} is used to map the annotation of all types of case markers in the CLE-UTB.}

\begin{table}[h]
\centering
\caption{The mapping of the CLE-UTB phrase labels on UD labels for nominal dependents.}
\label{tab-5}
\begin{center}
\begin{tabular}{@{}lll@{}}
\hline
CLE-UTB lables & UD-2.0 labels & UD Description \\
\thickhline
PP-G & nmod & Nominal modifier \\
CD & nummod & Numeric modifier \\
ADJP & amod & Adjectival modifier \\
JJ & amod & Adjectival modifier \\
S\textasciicircum{}ADJP & acl & Adjectival clause \\
S\textasciicircum{}NP & acl & Adjectival clause \\
S\textasciicircum{}PP & acl & Adjectival clause \\
DMP & det & Determiner \\
APNA & det & Determiner \\
FR & det & Determiner \\
OD & det & Determiner \\
PDM & det & Determiner \\
PRD & det & Determiner \\
PRF & det & Determiner \\
PRP & det & Determiner \\
PRR & det & Determiner \\
PRS & det & Determiner \\
Q & det & Determiner \\
QM & det & Determiner \\
PRE & case & Case marking \\
PSP & case & Case marking \\
PSP-G & case & Case marking \\
PSP-I & case & Case marking \\
PSP-A & case & Case marking \\
PSP-E & case & Case marking \\
PSP-SE & case & Case marking \\
\hline
\end{tabular}
\end{center}
\end{table}

\subsubsection{Other Dependency Relations}
\label{sec-3-2-4}
{\color{black}
Table~\ref{tab-6} presents the dependency labeling for coordination, multi-word expressions, punctuation, roots, and other unspecified dependencies. The conjunctions within noun, adjective, and quantifier phrases that are separated with commas are identified on the bases of POS tags and are marked with the {\tt conj} label. The coordinate conjunction tag CC is mapped on the {\tt cc} dependency label. The noun, adjective, and quantifier phrases associated with the internal case marker PSP-I are mapped on the {\tt fixed} label. The foreign fragment phrases mostly have a flat structure, therefore they are mapped on the {\tt flat} dependency label}. The compound nouns with more than one lexical items have rightmost word as a head-word and the non-head lexical items are mapped on the {\tt flat} label. An example of such noun phrases is \textit{mOm battI} `candle'. In Urdu, both \textit{mOm} `wax' and \textit{battI} `light' are nouns. In the head-word model, the noun \textit{battI} is the head and \textit{mOm} has a flat dependency relation on the head. 

\begin{table}[h]
\centering
\caption{The mapping of the CLE-UTB phrase labels on UD labels for dependency relations.}
\label{tab-6}
\begin{center}
\setlength\tabcolsep{15pt}
\begin{tabular}{@{}lll@{}}
\hline
CLE-UTB lables & UD-2.0 labels & UD description\\ 
\thickhline
NP* & conj & Conjunct \\
NN* & conj & Conjunct \\
ADJP* & conj & Conjunct \\
JJ* & conj & Conjunct \\
QP* & conj & Conjunct \\
Q* & conj & Conjunct \\
CC & cc & Coordinating conjunction \\
NP* & fixed & Fixed multiword expression \\
ADJP* & fixed & Fixed multiword expression \\
QP* & fixed & Fixed multiword expression \\
FF & flat & Flat multiword expression \\
NN & flat & Flat multiword expression \\
NP-POF & compound & Compound \\
ADJP-POF & compound & Compound \\
QP-POF & compound & Compound \\
VC-VALA & compound & Compound \\
LRB & punct & Punctuation \\
PU & punct & Punctuation \\
PU-C & punct & Punctuation \\
PU-E & punct & Punctuation \\
PU-P & punct & Punctuation \\
RRB & punct & Punctuation \\
VC & root & Root \\
FFP & dep & Unspecified dependency \\
PRT & dep & Unspecified dependency \\
\hline
\end{tabular}
\end{center}
* These dependency labels are achieved by using post-conversion rules.
\end{table}

The annotation of complex predicate structures in the treebank is represented by using the POF functional label. The functional label is generally attached with noun, adjective, and quantifier phrases. In the conversion process, phrases with POF label are mapped on the {\tt compound} label. The verbal construction with \textit{vAlA} particles are also marked as compounds. All types of punctuation marks are mapped on the {\tt punct} label. The heads of the verbal constructions are marked as {\tt root}. The verbal structure of the main clause at the highest level is marked as root. The {\tt dep} label is used to mark unspecified dependencies. The items in foreign fragment phrases have flat dependencies on heads, but their heads have an arc to the root of the sentence. We have marked such relations as unspecified dependencies. Similarly, particles are used as intensifiers and therefore marked with the {\tt dep} label during conversion.

\subsection{Post-Conversion Rules}
\label{sec-3-3}
For most of the phrase labels, dependency mapping is quite straight-forward as the phrase annotation scheme has functional labels that represent grammatical relations. Due to the flat annotation of the CLE-UTB, complex structures were mapped by implementing several post-conversion rules to generate an accurate {\color{black}DS} treebank. {\color{black}The first mapping issue is identified for the conversion of indirect objects. The annotation of the CLE-UTB does not annotate secondary objects but marks them as obliques by attaching OBL functional label. However, Urdu uses the accusative case marker \textit{kO} for secondary objects and recipients.} Similarly, non-finite clauses are annotated similar to finite clauses. These clauses are identified by their verbal heads. Furthermore, rules are derived for {\tt fixed} and {\tt conj} labels. Various rules are devised to make the generated treebank compatible with the Hindi-Urdu Treebank \cite{bhat2017hindi}. The details of the conversion rules are as follows.

\begin{enumerate}
{\color{black}
\item One way to represent secondary objects and recipients is the use of accusative/dative case marker \textit{kO} in Urdu. The annotation of the CLE-UTB annotates all case marking constructions with post-positional phrase (PP). The annotation marks secondary objects as obliques hence with the label PP-OBL. So the rule is that, if the label is the PP-OBL and the next lexical item is a clitic \textit{kO}, then map it on the {\tt iobj} dependency label.

\item There are nominal secondary objects that are annotated as nominal obliques. These constructions include certain types of pronouns to provide meanings of recipients or beneficiaries. A list of such pronouns with their meanings is shown in Table~\ref{tab-7}. If the phrase label is NP-OBL and the head of the phrase is any of these pronouns, then mark the dependency arc with the {\tt iobj} label.
}

\begin{table}[h]
\centering
{\color{black} 
\caption{List of pronouns marked as indirect objects.}
\label{tab-7}
\begin{tabular}{@{}lllll@{}}
\hline
Sr.\# & Pronouns & Romanized & Translation & Grammatical function\\
\thickhline

1 & \texturdu{مجھے} & \textit{mujHE} & To/for me & 1st person singular\\
2 & \texturdu{ہمیں} & \textit{hamEN} & To/for us & 1st person plural\\
3 & \texturdu{تجھے} & \textit{tujHE} & To/for you & 2nd person informal\\
4 & \texturdu{تمہیں} & \textit{tumEN} & To/for you & 2nd person polite/plural\\
5 & \texturdu{اِسے} & \textit{isE} & To/for him/her & Near, 3rd person singular\\
6 & \texturdu{اُسے} & \textit{usE} & To/for him/her & Far, 3rd person singular\\
7 & \texturdu{اِنہیں} & \textit{inhEN} & To/for them & Near, 3rd person plural\\
8 & \texturdu{اُنہیں} & \textit{unhEN} & To/for them & Far, 3rd person plural\\
9 & \texturdu{جسے} & \textit{jisE} & The one who & Relative pronoun, singular\\
10 & \texturdu{جنہیں} & \textit{jinhEN} & To/for those who & Relative pronoun, plural\\
11 & \texturdu{کسے} & \textit{kisE} & To/for whom & Interrogative pronoun\\
\hline
\end{tabular}
}
\end{table}

{\color{black}
\item The CLE-UTB annotates non-finite clauses and clausal objects similar to other clauses by using the `S' and S-OBJ labels. We map such constructions on the {\tt xcomp} dependency label. If a construction has any of these labels and their phrasal head is an infinitive verb (having the VBI POS tag), then map the construction with the {\tt xcomp} dependency label.

\item Some constructions contain an internal clitic as a connection between two nouns, adjectives, and quantifiers. For instance, the constituent \textit{kam-az-kam} `at least' has a clitic \textit{az} `from' in it. The clitic is tagged with the PSP-I POS tag. The constructions with the PSP-I tag are mapped on the {\tt fixed} dependency label.

\item The conjunctions between nouns, adjectives, and quantifiers are represented by using conjunction clitics and/or commas. A comma is used when there are more than two lexical items. We map such constructions on the {\tt conj} label if they use conjunctions with the CC or comma with the PU-C POS tag.

\item The HUTB annotates case markers that follow infinitive verbs, in a different manner. Such case markers are tagged with the {\tt mark} label rather than {\tt case}. We find such case markers and labeled them with the {\tt mark} dependency label to make our generated dependency treebank compatible with the HUTB.

\item The CLE-UTB annotated spacio-temporal nouns similar to common nouns. However, the HUTB uses the NST POS tag for such nouns and annotated them with the {\tt case} dependency label. Therefore, we compiled a complete list of such nouns from the HUTB and updated their labeling in the conversion process. If an NST noun appears after a case marker, then it is annotated with {\tt case} label and its head is the head of the preceding case marker. 

\item If two case markers appear together, then the head of next case marker is also the head of the previous case marker and they both are annotated with the {\tt case} dependency label.

\item The CLE-UTB annotates the spacio-temporal nouns by using a functional label ADJ. If a noun is not in NST list but has NP-ADJ as the original label, then it is assigned the {\tt obl} dependency label. If a token has NP-ADJ label but it is present in the NST list and the previous token is not a case marker, then it is marked as adverbial modifier by using the {\tt advmod} label.
}
\end{enumerate}

\begin{figure}[ht] \centering
\scalebox{0.44}{
\includegraphics[scale=0.44,clip, trim=0.9cm 0cm 4cm 3.5cm]{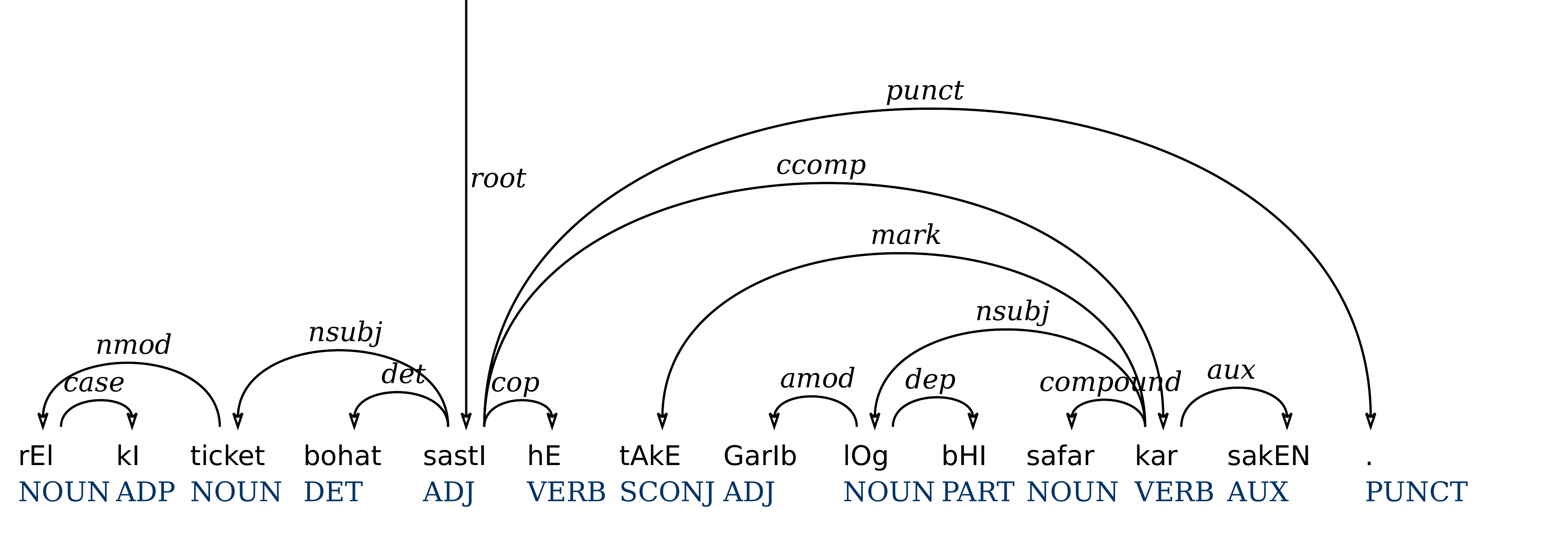}
}
\caption{Final dependency tree from the intermediate tree of Figure~\ref{fig-2} after label mappings and post-conversion rules.}
\label{fig-3}
\end{figure}

Figure~\ref{fig-3} presents the final dependency tree of the intermediate structure after applying label mappings and post-conversion rules.
The dependency arcs in Figure~\ref{fig-3} are quite similar to the arcs in Figure~\ref{fig-2} except for the copula construction. All phrase labels are replaced by the dependency labels. The copula construction is updated with respect to dependency labels. The predicate link is marked as {\tt root} of the sentence, and the copula verb \textit{hE} has a {\tt cop} dependency on the root. 
The generated treebank has 27 unique dependency labels that are given in Table~\ref{tab-8}.

\begin{table}[ht]
\centering
\caption{UD labels that are used by generated dependency treebank.}
\label{tab-8}
\begin{center}
\begin{tabular}{@{}llll@{}}
\hline
\multicolumn{4}{c}{Universal dependency labels} \\ 
\thickhline
			acl	& advcl & advmod & amod \\
			aux & case & cc & compound \\
			conj & cop & csubj & dep \\
			det & discourse & fixed & flat \\
			iobj & mark & nmod & nsubj\\
			nummod & obj & obl & punct \\
			root & vocative & xcomp & ---\\
\hline
\end{tabular}
\end{center}
\end{table}

\subsection{Conversion Evaluation}
\label{sec-3-4}
The evaluation is a mandatory task during the development of a computational data resource. Inter-annotator agreement scores are computed to represent the understanding of the annotators during the annotation process. In our case, the {\color{black}PS} treebank has already been evaluated using different methods. 
There was no reference corpus available to compute similarity or reference scoring for the generated dependency treebank. For this purpose, we obtained an existing dependency treebank from the Hindi-Urdu treebank project \cite{bhat2017hindi}. The project contains separate repositories for Hindi and Urdu. Each repository has training, test, and development sets. We took the first hundred sentences from the development set of the Urdu treebank repository and manually annotated them using the annotation guidelines of the CLE-UTB presented in \cite{ahmedmultilayered}. This reference corpus was then converted to the dependency structure and compared with its original annotation. This method may not completely evaluate the conversion process but provides an insight of the agreements for core arguments of the language.

The reference corpus has been evaluated by computing the labeled attachment score (LAS), the unlabeled attachment score (UAS), the label accuracy (LA), and Cohen's Kappa coefficient \cite{cohen1960coefficient}. Table~\ref{tab-9} shows the details of the evaluation scores.

\begin{table}[ht]
\centering
\caption{Inter-annotator agreement scores on reference corpus containing 100 sentences.}
\label{tab-9}
\begin{center}
\begin{tabular}{@{}llll@{}}
\hline
Category & UAS & LAS & LA \\
\thickhline
Dependency measures & 79.21 & 69.59 & 80.62\\
Kappa Coefficients & 0.7859 & 0.6942 & 0.7857\\
\hline
\end{tabular}
\end{center}
\end{table}

{\color{black}Evaluations are performed on the basis of tokens, their labels, and heads. The unlabeled attachment score shows the agreement of heads irrespective of their labels. However, the labeled attachment score shows the accuracy where both labels and heads are correct. The label accuracy shows the percentage of correct labels. Cohen's kappa coefficient computes the inter-rater agreement and its interpretation is shown in the Table~\ref{tab-10}. The agreement for heads and labels is quite acceptable as it is almost 0.80 showing strong agreement. However, the score for labeled attachment is near 0.70 which represents moderate agreement.}

\begin{table}[ht]
\centering
\caption{Interpretation of Cohen's Kappa coefficient values.}
\label{tab-10}
\begin{center}
\begin{tabular}{@{}ll@{}}
\hline
Kappa value & Agreement level \\
\thickhline
0 - 0.20 & None\\
0.21 - 0.39 & Minimal \\
0.40 - 0.59& Weak \\
0.60 - 0.79& Moderate\\
0.80 - 0.90& Strong\\
0.91 and above & Almost Perfect \\
\hline
\end{tabular}
\end{center}
\end{table}

The annotation of both treebanks is different as the Urdu treebank from the HUTB project has been annotated manually. On the other hand, our dependency treebank is generated from a {\color{black}PS} treebank by utilizing functional labels to represent dependency relations. These labels are limited to represent all types of dependency relations. However, the generated dependency has the representation of major types of core and non-core arguments.

\section{Sequential Representations}
\label{sec-4}
The parsing task can be simplified by transforming the hierarchical representation into sequence labeling. Both phrase and dependency structures can be converted to sequence labels and back to the hierarchical structure accurately. Sections~\ref{sec-4-1} and \ref{sec-4-2} present the transformation of PS and DS trees into sequence labels.

\subsection{PS Tree Representation}
\label{sec-4-1}
\begin{figure}[ht]
\centering 
\scalebox{0.59}{
\includegraphics[scale=0.39,clip, trim=0.5cm 0cm 2.5cm 1cm]{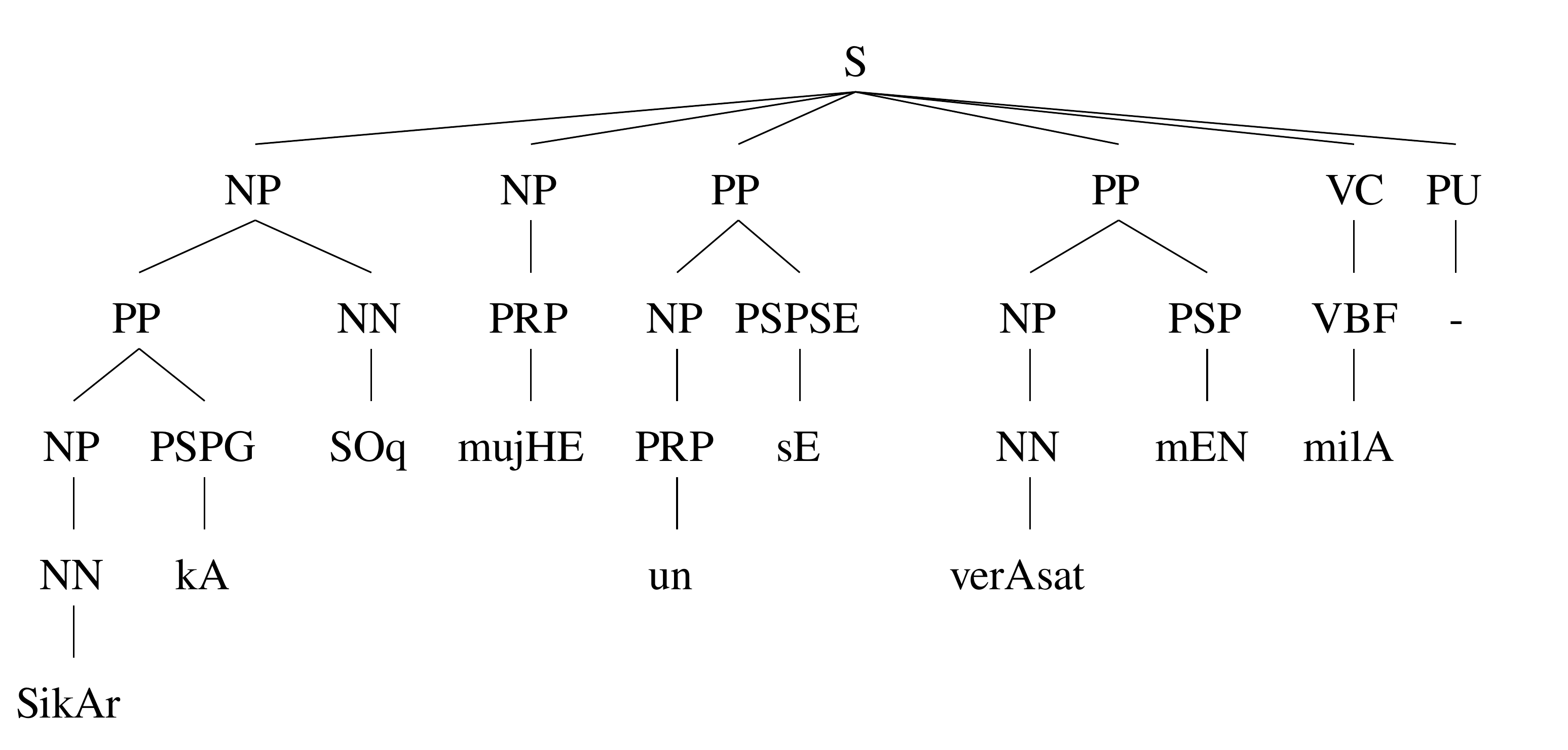}
 } 	\\
\setlength\tabcolsep{3pt}
\begin{tabular}{lllll} 
SikAr=kA & SOq & mujHE & un=sE & varAsat=mEN \\
hunting.Sg=Gen & passion.Sg.Nom & Pron.1.Sg & pron.3.Pl=Abl & inheritance.Sg=Loc \\
milA & & & & \\
find.Past.3.Sg & & & & \\
\multicolumn{5}{l}{`I inherited the passion of hunting from him'.}
\end{tabular} 
\caption{A sample {\color{black}PS} parse tree along with sequence labels. }
\label{fig-4}
\end{figure}

For the conversion of PS trees to sequential format, we use a baseline transformation approach as discussed in \cite{carlos2018Constituent}. The label output is in CoNLL format containing tokens, POS tags, and tree labels. The treebank labels consist of an integer and a tag. The integer, which is part of a label, represents the total number of common ancestors. The phrase label, attached with the integer, represents the common label between successive tokens at the lowest level in the parse tree. However, unary branches are handled separately by attaching them with POS tags. Figure~\ref{fig-4} shows an Urdu PS parse tree followed by the proposed linear labeling (\ref{g-9}) compared to the relative and absolute labeling presented in \cite{carlos2018Constituent}.

{\color{black}
\ex
\begingl 
\label{g-9}
\gla Tokens: SikAr kA SOq mujHE un sE varAsat mEN milA -//
\glb Relative~\cite{carlos2018Constituent}: 3\_PP -1\_NP -1\_S 0\_S 1\_PP -1\_S 1\_PP -1\_S 0\_S NONE//
\glb Absolute~\cite{carlos2018Constituent}: 3\_PP 2\_NP 1\_S 1\_S 2\_PP 1\_S 2\_PP 1\_S 1\_S NONE//
\glb Proposed(our's): PP 2\_NP 1\_S 1\_S PP 1\_S PP 1\_S 1\_S NONE//
\glft ~`I inherited the passion of hunting from him'.//
\endgl
\xe
}

A simple (absolute) linear encoding produces a label using the common number of phrase labels between $w_i$ and $w_i+1$. However, it may generate a large number of labels. The second encoding method proposed by \cite{carlos2018Constituent} uses the difference of common ancestors from the next $w_t+1$ and previous token $w_t-1$. For example, the token \textit{kA} (genitive case marker) has a relative scale label -1\_NP. The token \textit{kA} has two common ancestors, NP and S, with the next token \textit{SOq} `passion' and three common ancestors, PP, NP and S, with the previous token \textit{SikAr} `hunting'. Therefore, the difference is -1 and the lowest common ancestor label with the next token is NP which produces the linear label -1\_NP. This method is referred as relative scale and generates remarkably less number of labels as compared to absolute scale encoding. The unary branch labels are produced separately at the first step of the labeling that are further combined with POS tags for the purpose of conversion back to brackets (phrase structure). Therefore, the experiments are performed in two passes. The first pass predicts unary labels and the second pass learns the relative phrase labels. The output is converted back to parse trees to perform evaluations.

{\color{black}
The analysis and experiments of relative and absolute labeling were performed for the English Penn Treebank. The annotation of the CLE-UTB has a relatively flat structure hence with less number of labels. We propose a labeling scheme that has the ability to generate even fewer labels. The average entropy of our proposed labeling is also lower than the relative labeling. The proposed label set produces better parsing scores for the CLE-UTB and outperformed the relative labeling scheme. Table~\ref{tab-12} presents the comparison of both labeling schemes.}

\begin{table}[h]
\centering
\caption{Comparison of the proposed sequential labeling with the labeling of \cite{carlos2018Constituent} with respect to the CLE-UTB.}
\label{tab-12}
\begin{center}
\begin{tabular}{@{}lll@{}}
\hline
Sequence labeling & \# of labels & Entropy (average) \\
\thickhline
Relative \cite{carlos2018Constituent} & 180 & 3.15\\
Proposed(our's)& 143 & 2.65\\
\hline
\end{tabular}
\end{center}
\end{table}

Table~\ref{tab-12} shows the number of labels and the average entropy for the entire treebank using relative and proposed labeling schemes. The entropy values depict the randomness of labels for a sentence. $X$ is the random variable given that has possible outcomes $x_1, x_2,...,x_n$ with probability $P(x_1),P(x_2),...,P(x_n)$, the entropy of the variable $X$ is represented by equation~\ref{eq-0}.

\begin{equation}
\label{eq-0}
H\left( X\right) =-\sum ^{n}_{i=1}P\left( x_{i}\right) \log P\left( x_{i}\right)
\end{equation}

The entropy value of the proposed labeling is less than the relative labeling, showing that the proposed labeling has less randomness and hence has a higher learning capability in the training process. The proposed labeling is based on the absolute labeling approach that labels a common number of phrase labels of a token and its next token. The label further concatenates the lowest common phrase with it. Similarly to relative labeling, a label consists of a number followed by a phrase label. 

{\color{black}
Absolute labels were analyzed against the annotation of the CLE-UTB to achieve predictable patterns. There were several patterns due to the simplified and flatter syntactic structure of the PS treebank. We analyzed them to reduce the number of labels. The number of common phrase labels in a PP phrase is always one higher than the number of next labels. In absolute labeling, the first label is 3\_PP and the second label is 2\_NP. Similarly, fifth and seventh tokens have 2\_PP label and their next label is 1\_S. This sequence is found throughout the treebank. We dropped the integer from all PP labels that in result reduced the size of the label set. A similar pattern was observed for the PP-G labels and the size of the label set was further reduced.}

Another pattern was observed for VC labels. The number of common phrase labels is always one higher than the common phrase labels of the previous label. The integer part was removed from all the VC labels. The demonstrative phrase (DMP) showed a pattern that was similar to post-positional labels. The DMP labels were also simplified to reduce the labels by removing integers from them. These integers were recovered after an output was obtained from the parser. There were some patterns with NP labels, but they rather decreased the parsing scores; therefore, we utilized the four aforementioned patterns. The proposed labeling reduced the size of the tag set to 143 labels compared to 180 labels from the relative labeling.

{\color{black}
\subsection{DS Tree Representation}
\label{sec-4-2}

DS parse trees are represented in graphical as well as CoNLL format. The graphical form uses arcs to represent dependencies which are further labeled with dependency labels as demonstrated by Figure~\ref{fig-5}. On the other hand, CoNLL format demonstrates the structure of a sentence in columns. Each token has a unique number from 1 to $n$, if there are $n$ tokens in the sentence. Heads are shown by adding their token numbers and unique dependency label is assigned to each token. The CoNLL format makes it compatible to have sequential labels against tokens. A sequence labeling for the {\color{black}DS} has been introduced in \cite{spoustova2010dependency}. Similarly, further labeling schemes are presented in \cite{strzyz2019viable}. They presented four sequence labeling methods and compared the dependency parsing results. There labeling formats include: naive positional, relative positional, relative POS-based and bracketing-based labels. The relative POS-based label method outperformed the other three encodings with respect to dependency parsing.

\begin{figure}[h]
\centering 
\scalebox{0.45}{
\includegraphics[scale=0.45,clip, trim=0.5cm 0cm 3.5cm 14cm]{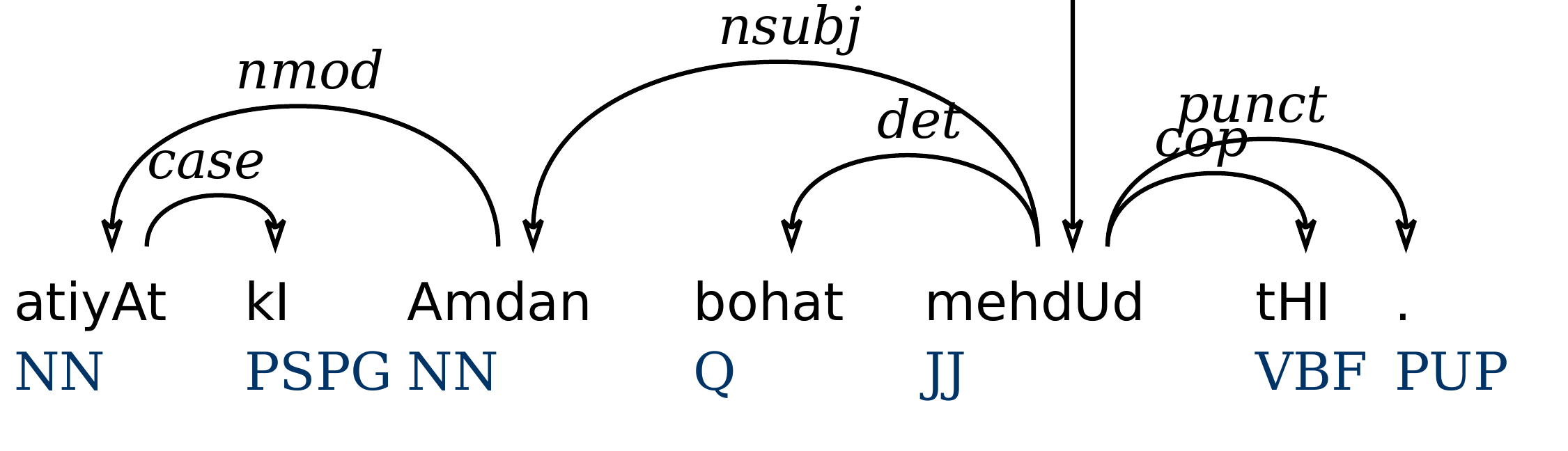}
 } 	\\
\setlength\tabcolsep{8.5pt}
\begin{tabular}{lllll} 
atiyAt=ki & Amdan & bohat & mehdUd & tHI \\
donation.Pl=Gen & income.Sg.Nom & very & limited & be.Perf.3.F.Sg \\
\multicolumn{5}{l}{`Donation income was very limited'.}
\end{tabular} \\ 
\caption{A sample {\color{black}DS} parse tree along with its sequence labels. }
\label{fig-5}
\end{figure}

{\color{black}
\ex
\begingl 
\label{g-10}
\gla Tokens: atiyAt ki Amdan bohat mehdUd tHI .//
\glb Relative~\cite{strzyz2019viable,spoustova2010dependency}: +1,NN,nmod -1,NN,case +1,JJ,nsubj +1,JJ,det -1,ROOT,root -1,JJ,cop .//
\glft ~`Donation income was very limited'.//
\endgl
\xe
}

Figure~\ref{fig-5} shows a sample parse tree containing seven tokens followed by a relative POS-based encoding (\ref{g-10}). The encoding is three tuple $(x_i,p,l_i)$. $x_i$ represents the relative position of the head with respect to the POS tag of the head token. It further uses the symbols $+/-$ to represent the direction to find a head. For example, the first token \textit{atiyAt} `donations' has the label {\tt +1,NN,nmod}. Its head is \textit{Amdan} `income' that is on the right-hand side of \textit{atiyAt} and has a POS tag NN. The label contains a $+1$ to show that the first token from right which has POS tag NN is the head of \textit{atiyAt}. If the head is on the left side of the token $i$, then it will be represented by $-$ symbol. We have used this encoding for both single-task and multi-task parsing experiments.

However, the naive positional encoding is two tuple $(x_i,l_i)$, where $x_i$ is the head token number and $l_i$ is the label of token $i$. The relative positional encoding is also two tuple $(x_i,l_i)$, but it generates the relative position of a head for each token. It uses symbols $(+/-)$ to depict position of heads. Relative POS-based encoding is three tuple $(x_i,p,l_i)$. It adds the relative position of a head token with respect to head's POS tag. 
Finally, the bracketing-based encoding is also two tuple $(x_i,l_i)$. It contains bracket symbols to represent incoming and outgoing arcs for a token $i$ along with its label $l_i$. We employed the relative POS-based encoding for dependency parsing that has produced state-of-the-art dependency parsing results for the CLE-UTB.
}

\section{Multi-Task Parsing}
\label{sec-5}
{\color{black}The proposed multi-task parsing model is based on {\color{black}Long Short-Term Memory} (LSTM) networks. We propose two main models with different features. The first model is applied as single-task learning to perform the constituency and dependency parsing independently. The second model performs multi-task learning by incorporating cross-representations and achieves improved parsing results. The following sections describe both proposed models and transfer learning paradigms.}

\subsection{Model}
\label{sec-5-1}
\begin{figure}[ht] \centering
\includegraphics[scale=0.50]{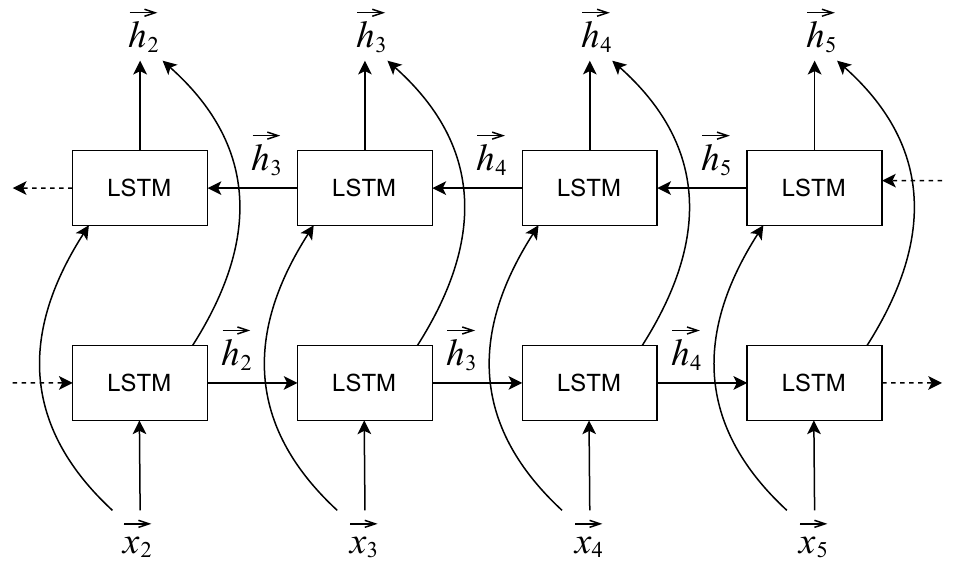}
\caption{Bi-directional {\color{black}Long Short-Term Memory} model for sequence labeling.}
\label{fig-6}
\end{figure}

{\color{black}Recurrent neural networks (RNNs) are able to provide a learning framework to model word sequences using the entire context. Our parsing models are built on bidirectional {\color{black}Long Short-Term Memory} (BLSTM) networks, a variant of conventional RNNs. The parse trees are converted to linear structures along with tree labels and POS tags to make them compatible with the BLSTM structure. Figure~\ref{fig-6} shows the model architecture having two LSTM layers, one in the forward direction and the other in the backward direction.}

\subsubsection{Single-Task Learning Model}
\label{sec-5-1-1}
\begin{figure}[ht] \centering
\scalebox{0.85}{
\includegraphics{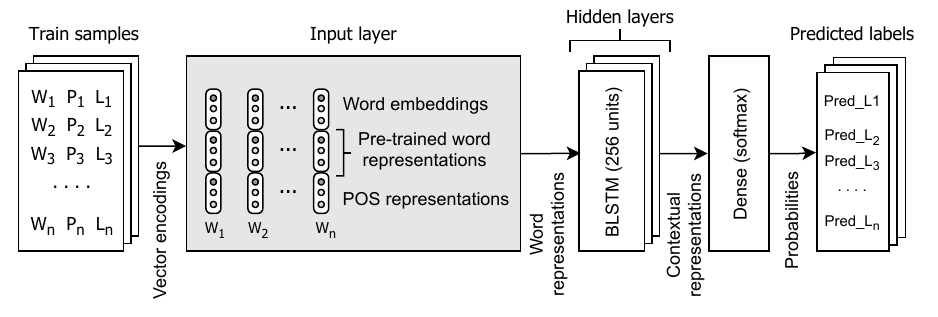}
}
\caption{Architecture of the single-task learning model.}
\label{fig-8}
\end{figure}

The single-task learning model performs constituency and dependency parsing independently. Data representation is quite similar for both types of syntactic structures; therefore, the same parsing model is trained. Figure~\ref{fig-8} shows the architecture of the single-task learning-based parsing model. 

{\color{black}The training set contains tokens, POS tags, and sequence labels in CoNLL format that are fed to the input layer in the form of vector encodings. Vector encodings consist of word embeddings, POS embeddings, and pre-trained word representations to perform transfer learning. At the input layer, we performed experiments with several different features to demonstrate the improvements in the parsing results. The features include character embeddings, POS tags and pre-trained word embeddings. All vector representations are concatenated to form a single vector for each token and are provided to hidden LSTM layers. Each hidden LSTM layer is set to have 256 hidden units. The contextual representations from hidden layers are then given to a dense output layer with $softmax$ activation to perform the multi-class classification. The $RSMprop$ optimizer is utilized with a learning rate of $0.001$. The batch size is set to 32. The model was trained for 25 epochs. The parsing experiments are performed on a scientific computer cluster (https://www.scc.uni-konstanz.de).}

\subsubsection{Multi-Task Learning Model}
\label{sec-5-1-2}
\begin{figure}[ht] \centering
\scalebox{0.95}{
\includegraphics{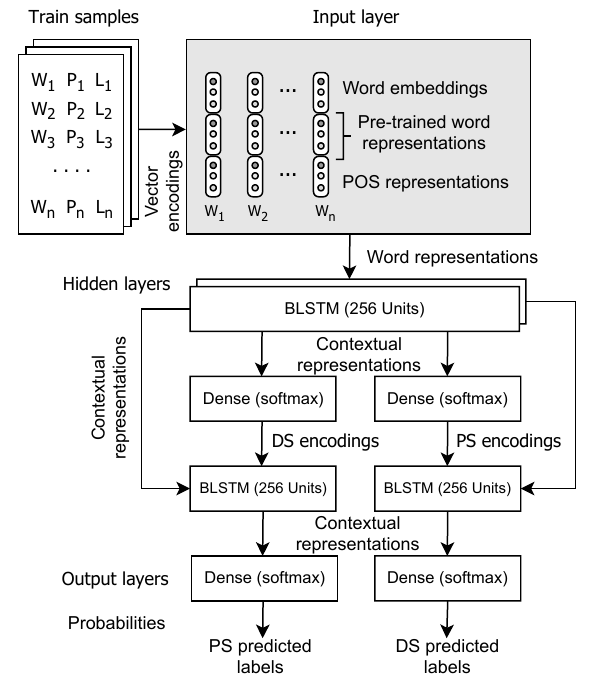}
}
\caption{Multi-task learning based model architecture.}
\label{fig-9}
\end{figure}

The multi-task learning model utilizes cross-representations to perform both types of parsing. The learned {\color{black}PS} encodings are used for the training of the {\color{black}DS} and vice versa. Cross-representations are quite helpful to improve the results for both constituency and dependency parsing. Figure~\ref{fig-9} shows the architecture of the multi-task learning model. 

The input layer of both single-task and multi-task learning is similar. They both use word embeddings, POS embeddings, and pre-trained word representations. Pre-trained word representations enable transfer learning that is detailed in Section~\ref{sec-5-2}. In the multi-task learning model, we use two BLSTM hidden layers, each having 256 hidden units. The contextual representations from hidden layers are provided to two difference $softmax$ layers for multi-class classification to achieve phrase and dependency encodings separately. These encodings are further fed to a single BLSTM layer along with contextual representations from the previous two BLSTM hidden layers. PS encodings are used to train DS representations, and DS encodings are used to train PS representations. Finally, two output $softmax$ layers are used to predict the PS and DS labels. The $RSMprop$ optimizer is used with a learning rate of $0.001$ and a batch size of 32 samples is set for the training of multi-task model and the model is trained for 25 epochs.

\subsection{Transfer Learning}
\label{sec-5-2}
It is costly to create large annotated data sets for a language. Therefore, transfer learning is a suitable approach to resolve the issue of data sparsity by training word representations on large unannotated corpora. It also addresses the problem of insufficient vocabulary. To train two parsing models for Urdu, we trained word representations on a plain text corpus containing 220 million Urdu tokens collected from the Word Wide Web. The corpus has text from a number of different text genres like news, business, culture, health, sports, etc. We trained context-free word embeddings and contextualized word representations in our experiments.

\subsubsection{Context-Free Word Representations}
\label{sec-5-2-1}
Context-free word representations generate a single feature vector for each token in the vocabulary. In the results, a single meaning is associated with each token. However, these representations are capable of learning the syntactic and semantic details of words for a language. Well-known word representations are Word2Vec\cite{mikolov2013efficient} and GloVe\cite{pennington2014glove} embeddings among others. Both embeddings are implemented using different algorithms. Word2Vec is trained on the bases of co-occurrence of consecutive words, while GloVe embeddings are based on word-to-word co-occurrences. Word2Vec embeddings are further trained by implementing two techniques, which are the continuous bag-of-words (CBoW) and skip-gram model. The CBoW model learns the surrounding tokens of the input layer and predicts the pivotal word. On the other hand, the skip-gram is a feed-forward neural network model that learns a token from the input layer and predicts the surrounding tokens in a selected window. The size of the widow can be adjusted to enhance the contextual information. In both algorithms, a feature vector is obtained from the hidden layer. We trained Word2Vec embeddings using a skip-gram model for transfer learning in our parsing experiment. The vocabulary contained 125,000 tokens with 100 dimensions for each token. A window size of five was selected for training. Word2Vec embeddings are trained using gensim library (https://pypi.org/project/gensim). Section~\ref{sec-6} presents the results of the parsing models using Word2Vec word embeddings.

\subsubsection{Contextualized Word Representations}
\label{sec-5-2-2}
\begin{figure}[ht] \centering
\includegraphics[scale=0.80]{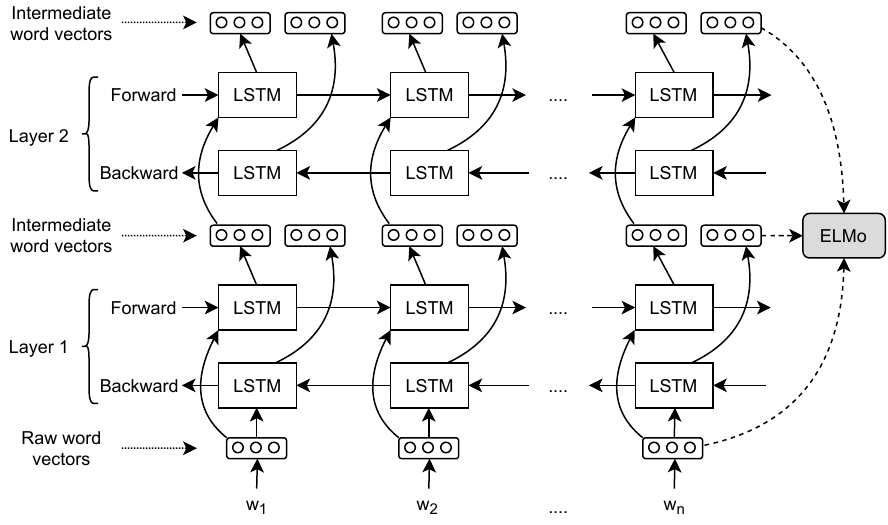}
\caption{Architecture of ELMo word representation model.}
\label{fig-10}
\end{figure}

In natural languages, different meanings are generally associated with words according to their context. The contextual information is quite useful for performing a sequence labeling task. Therefore, we also trained deep contextualized word representations (ELMo) (https://allennlp.org/elmo) \cite{peters2018deep} for the purpose of transfer learning. ELMo (Embeddings from Language Models) representation is helpful to achieve state-of-the-art results for many natural language processing tasks. These representations learn word vectors from deep bidirectional language models by employing character-based convolution neural networks (CNNs), which are further used to generate vectors for out-of-vocabulary (OOV) tokens as well. Character-based embeddings are also useful for learning the morphological insight of a language. The feature vectors for all tokens of a sentence are computed from pre-trained ELMo weights. These vectors use the contextual information of each token and hence a token may have different vectors according to its semantics in a sentence. Figure~\ref{fig-10} shows the model architecture of ELMo representations.

The ELMo word vectors are computed on the top of the two-layer bidirectional language model. Each layer has two passes; the first is the forward pass and the second is the backward pass. The architecture, shown in Figure~\ref{fig-10}, uses a character-level CNN network and represents a sentence in raw word vectors at the input layer. The forward pass contains the information of a token and its context before that certain token, while the backward pass contains the information of the token and the context after that token. This contextual information forms the intermediate word vectors which are further provided to the next layer of bidirectional language model. The final representation is achieved by weighted sum of the raw vectors and two intermediate vectors for each token. However, three types of vector can be produced at each layer that are the input layer, first layer, and second layer. The final layer produces the contextual word representations that contain complete contextual information. The contextualization is useful for improving sequence labeling and to perform word sense disambiguation. The same corpus containing 220 million tokens was used to train the ELMo embeddings. We utilized word vectors produced by the last layer of the bidirectional language model in the parsing experiments.

\section{Results and Discussions}
\label{sec-6}
This section presents the parsing results and their interpretation. The CLE-UTB contains 7,854 sentences with 148,475 tokens. {\color{black}The corpus is further divided into train and test sets. The train set contains 6,881 sentences and 130,105 tokens, while the test set contains 973 sentences with 18,470 tokens. We created the test set in a way that includes the sentences from all genres of text in the corpus}. Domain-wise details of the corpus are given in \cite{ehsan2021development}.

\subsection{Dataset}
\label{6-1}
\begin{figure}[ht] 
\centering
\scalebox{0.75}{
\includegraphics{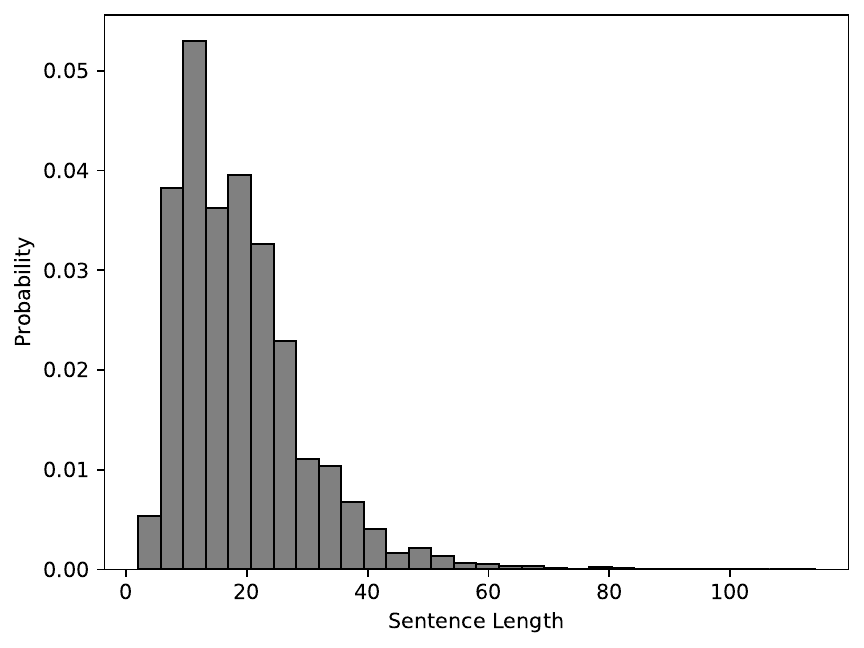} 
}
\caption{Histogram for sentence lengths in the dataset.}
\label{fig-12}
\end{figure}

In this section, brief statistics of the dataset are presented. {\color{black}Figure~\ref{fig-12} shows a histogram prepared with respect to sentence lengths. The x-axis shows the length groups and the y-axis presents the percentage of the sentences according to the number of tokens in them. The corpus has the highest presence of sentences with length approximately equal to ten. However, the average sentence length of the corpus is 18.9 tokens. The corpus also has sufficient coverage of longer sentences, as shown in the diagram. There are many sentences that have more than one hundred tokens in them. The histogram shows that the treebank has a natural coverage of the text with sentences of various lengths.}


Table~\ref{tab-12-2} shows the frequency of the phrase and functional labels in the constituency treebank. The treebank has 11 phrase labels and 10 functional labels. The frequencies in the table are arranged in decreasing order. The NP phrase has most occurrences, and the PREP phrase has lowest appearances. On the other hand, SUBJ functional label appears most of the time that marks the subject role, and the INJ label has lowest frequency that is used to annotate interjections. {\color{black}In general, the treebank has sufficient coverage of the core arguments of the grammar of the language.}

\begin{table}[ht]
\centering
\caption{Frequencies of occurrence for phrase and function labels in the {\color{black}PS} treebank.}
\label{tab-12-2}
\begin{center}
\begin{tabular}{@{}lllll@{}}
\hline
Sr.\# & Phrase labels & Frequency & Function labels & Frequency\\
\thickhline
1 & NP & 48,359 & SUBJ & 10,417\\
2 & PP & 21,416 & G & 8,923\\
3 & S & 20,791 & POF & 5,293\\
4 & VC & 17,530 & OBJ & 4,064\\
5 & SBAR & 4,741 & ADJ & 3,080\\
6 & ADJP & 3,331 & PDL & 3,040\\
7 & ADVP & 2,114 & OBL & 2,321\\
8 & QP & 1,136 & VOC & 257\\
9 & FFP & 138 & VALA & 213\\
10 & DMP & 57 & INJ & 98\\
11 & PREP & 33 & --- & ---\\
\hline
\end{tabular}
\end{center}
\end{table}

Table~\ref{tab-12-3} presents the frequencies of the dependency labels that appear in the generated {\color{black}DS} treebank. In total, 27 unique dependency labels are generated in the conversion process. The {\tt case} label has the highest frequency of 22,630 depicting a strong case system of the language. Urdu uses separate clitics representing cases hence the {\color{black}DS} has  additional dependency labels for each clitic. On the other hand, the {\tt fixed} label has the lowest frequency. The {\tt fixed} label has been used for internal case markers that have quite a few occurrences in the corpus.

\begin{table}[h]
\centering
\caption{Frequencies of occurrence for dependency labels in the {\color{black}DS} treebank.}
\label{tab-12-3}
\begin{center}
\begin{tabular}{@{}llllllll@{}}
\hline
Dep. label & Freq. & Dep. label & Freq. & Dep. label & Freq. & Dep. label & Freq.\\
\thickhline
case & 22,630 & root & 7,854 & acl & 3,931 & iobj & 225\\
nmod & 13,307 & amod & 7,297 & advmod & 3,727 & discourse & 203\\
punct & 13,111 & det & 5,677 & cop & 3,011 & xcomp & 129\\
aux & 11,532 & conj & 5,387 & nummod & 2,704 & flat & 126\\
obl & 11,294 & cc & 4,516 & dep & 2,180 & csubj & 63\\
nsubj & 10,354 & mark & 4,025 & advcl & 1,775 & fixed & 8\\
compound & 9,250 & obj & 4,000 & vocative & 255 & --- & --- \\
\hline
\end{tabular}
\end{center}
\end{table}

{\color{black}
By analyzing the core arguments and labels in both treebanks, it can be concluded that the conversion is quite accurate. For example, the constituency treebank has 10,417 subjects that generate 10,354 nominal subjects ({\tt nsubj}) and 63 clausal subjects ({\tt csubj}) for the {\color{black}DS}. Similarly, the constituency treebnank contains 4,064 objects that are marked with an OBJ functional label. In the generated dependency treebank, there are 4,000 direct objects marked with the {\tt obj} label and there are 64 clausal objects in the constituency structure that are further mapped on the {\tt xcomp} dependency label according to the third rule in Section~\ref{sec-3-3}. However, there is a significant difference in the appearance of oblique constructions. The generated treebank contains over 11,000 occurrences of the {\tt obl} label, but the constituency treebank has 2,321. The reasons for this inflation are ADJ annotations in the constituency treebank and a list of spacio-temporal nouns extracted from HUTB. According to the ninth post-conversion rule, all phrases having phrases with the ADJ functional label or having spacio-temporal nouns, are mapped on the {\tt obl} label. This additional mapping significantly increased the frequency of {\tt obl} in the generated treebank. 
}

\subsection{Evaluation Metrics}
\label{6-2}

To evaluate the constituency parsers, labeled recall (LR), labeled precision (LP) and labeled F-score (F1) are used. Labeled measures consider that a certain constituent is correct if tokens in constituents along with their labels are correct. {\color{black}Equations~\ref{eq-9}, \ref{eq-10} and \ref{eq-11} present the formulas for the evaluation measures of the constituency parser.}

\begin{equation}
\label{eq-9}
Labeled\;Recall\;(LR) = \dfrac{\#\;Correct\;constituents\;in\;candidate\;set}{\#\; Constituents\;in\;reference\;set}
\end{equation}

\begin{equation}
\label{eq-10}
Labeled\;Precision\;(LP) = \dfrac{\#\;Correct\;constituents\;in\;candidate\;set}{\#\; Constituents\;in\;candidate\;set}
\end{equation}

\begin{equation}
\label{eq-11}
Labeled\;F-score\;(F_1) = \dfrac{2.LR.LP}{LR + LP}
\end{equation}

The {\color{black}DS} {\color{black}contains the information of the heads and their labels at the token level; therefore, its evaluation measures are different compared to constituency parsing.} The metrics used are the unlabeled attachment score (UAS), labeled attachment score (LAS), and label accuracy (LA). The unlabeled attachment score is the percentage of correct heads with respect to all tokens in the test set. {\color{black}The labeled attachment score is the percentage of correct heads along with correct labels, and the label accuracy is the accuracy of correct labels irrespective of the heads.} The unlabeled attachment scores and label accuracy are usually higher than the labeled attachment scores. Equations~\ref{eq-12}, \ref{eq-13}, \ref{eq-14} present general formulas for dependency evaluation metrics. 

\begin{equation}
\label{eq-12}
Unlabeled\;Attachment\;Score\;(UAS) = \dfrac{\#\;Tokens\;with\;correct\;heads} {\#\;Tokens\;in\;candidate\;set}
\end{equation}

\begin{equation}
\label{eq-13}
Labeled\;Attachment\;Score\;(LAS) = \dfrac{\#\;Tokens\;with\;correct\;heads\;and \;labels} {\#\;Labels\;in\;candidate\;set}
\end{equation}

\begin{equation}
\label{eq-14}
Label\;Accuracy\;(LA) = \dfrac{\#\;Tokens\;with\;correct\;labels} {\#\;Labels\;in\;candidate\;set}
\end{equation}

\subsection{Results}
\label{6-3}

In the experiments, BLSTM-based models are trained along with various features that include character embedding, POS tags, Word2Vec embeddings, and ELMo representations. {\color{black}Table~\ref{tab-13} shows the results of the constituency parsing based on single-task learning.} The experiments are conducted with and without functional labels. The standalone BLSTM-based models provide baseline F-scores of 78.98 and 81.41 with and without functional labels. The F-scores demonstrate a subtle improvement when trained by utilizing character embeddings. {\color{black}Using POS tag encodings along with word embeddings, the results improve by five points when trained without functional labels and by about four points for functional labels}.

\begin{table}[ht]
\small
\centering
\caption{Constituency parsing results by using single-task learning paradigm.}
\label{tab-13}
\begin{center}
\setlength\tabcolsep{5pt}
\begin{tabular}{@{}llllllll@{}}
\hline
Models \& Features & \multicolumn{3}{c}{No functional tags} & \multicolumn{4}{c}{Functional tags}\\
& $Rec.$ & $Pre.$ &$F_1$ & $Rec.$ & $Pre.$ & $F_1$ & F. acc. \\
\thickhline
BLSTM & 80.94 & 81.89 & 81.41 & 75.47 & 82.83 & 78.98 &73.72\%\\
BLSTM+ch-emb & 80.95 & 82.11 & 81.53 & 76.1 & 83.49 & 79.62 &74.36\%\\
BLSTM+ch-emb+pos & 86.04 & 87.24 & 86.64 & 80.37 & 87.29 & 83.69 &78.43\%\\
BLSTM+ch-emb+pos+w2v & 86.75 & 88.37 & 87.55 & 81.18 & 88.23 & 84.56 &79.30\%\\
BLSTM+pos+elmo & 90.07 & 91.68 & \textbf{90.86} & 89.81& 91.47& \textbf{90.63} & \textbf{84.37}\%\\
BLSTM+pos+elmo+w2v & 89.88 & 90.93 & 90.40 &88.88&90.65&89.76& 84.50\% \\
CNN-BLSTM+pos & 86.25 & 88.14 & 87.18 & 85.79 & 87.45 & 86.61& 82.65\% 
\\
CNN-BLSTM+pos+w2v &88.59 &89.21 &88.90 &85.59 &87.20 &86.39 &82.40\% \\
CNN-BLSTM+pos+elmo &90.39 &91.49 & \textbf{90.94} &90.15 &91.17 & \textbf{90.66} & \textbf{84.72}\% \\
CNN-BLSTM+pos+elmo+w2v &90.50 &91.22 &90.86 &89.68 &90.70 &90.19 &83.85\% \\
\hline
\end{tabular}
\end{center}
\end{table}

\begin{table}[ht]
\small
\centering
\caption{Phrase-wise constituency parsing results by using single-task learning paradigm.}
\label{tab-13-2}
\begin{center}
\begin{tabular}{@{}llllllll@{}}
\hline
Sr.\# & Phrase labels & \multicolumn{3}{c}{No functional tags} & \multicolumn{3}{c}{Functional tags}\\
&& $Rec.$ & $Pre.$ &$F_1$ & $Rec.$ & $Pre.$ & $F_1$ \\
\thickhline
1& NP & 89.07 & 89.60 & 89.34 & 89.60 & 89.20 & 89.40\\
2& PP & 90.66 & 92.82 & 91.73 & 90.81 & 93.08 & 91.93\\
3& S & 87.19 & 88.47 & 87.82 & 84.40 & 87.67 & 86.01\\
4& VC & 98.19 & 97.82 & 98.00 & 98.42 & 97.83 & 98.12\\
5& SBAR & 86.11 & 85.96 & 86.03 & 82.33 & 83.48 & 82.90\\
6& ADJP & 75.70 & 80.22 & 77.89 & 75.70 & 78.10 & 76.88\\
7& ADVP & 89.45 & 91.38 & 90.41 & 89.87 & 88.75 & 89.31\\
8& QP & 72.11 & 79.10 & 75.44 & 65.99 & 78.86 & 71.85\\
9& FFP & 66.67 & 87.50 & 75.68 & 71.43 & 78.95 & 75.00\\
10& DMP & 0.00 & nan & nan & 0.00 & 0.00 & nan\\
11& PREP & 0.00 & 0.00 & nan & 0.00 & nan & nan\\
\hline
\end{tabular}
\end{center}
\end{table}

\begin{table}[ht]
\small
\centering
\caption{Label-wise function label F-scores by using single-task learning paradigm.}
\label{tab-13-3}
\begin{center}
\setlength\tabcolsep{10pt}
\begin{tabular}{@{}lllll@{}}
\hline
Sr.\# &Function labels & $Rec.$ & $Pre.$ & $F_1$ \\
\thickhline
1& SUBJ & 90.01 & 86.81 & 88.38\\
2& G & 99.44 & 99.16 & 99.30\\
3& POF & 83.64 & 83.26 & 83.45\\
4& OBJ & 63.17 & 66.75 & 64.91\\
5& ADJ & 84.22 & 85.37 & 84.79\\
6& PDL & 78.61 & 85.0 & 81.68\\
7& OBL & 41.31 & 47.56 & 44.21\\
8& VOC & 68.97 & 86.96 & 76.92\\
9& VALA & 95.24 & 86.96 & 90.91\\
10& INJ & 100.0 & 100.0 & 100.0\\
\hline 
\end{tabular}
\end{center}
\end{table}

We perform transfer learning by training context-free Word2Vec and contextualized ELMo word representations. In the first step, we used Word2Vec embeddings along with character encodings and POS tags. It improved the results by about 0.9 points for both cases. The ELMo representations provide character-based contextualized word embeddings, therefore, we applied ELMo representations without utilizing raw character encodings. The ELMo embeddings provide substantial improvements in the results. Without functional labels, it produces an F-score of 90.86 and an F-score of 90.63 using functional labels for the CLE-UTB. These F-scores are quite promising for a {\color{black}morphologically rich} language. The F-scores show that ELMo embeddings are quite capable of learning the morphology and syntax of the language. We further trained the single-task model by combining ELMo and Word2Vec word embeddings, but scores started to decrease. 

The second portion of Table~\ref{tab-13}, presents F-scores for a CNN-BLSTM hybrid model along with the features. The hybrid model outperforms the BLSTM models and follows a similar trend with higher F-scores. The CNN-BLSTM model that uses POS tags and ELMo embeddings provides F-scores of 90.66 and 90.94 when experimented with and without functional labels, respectively. On the other hand, the accuracy of functional labels has also increased to 84.72\%. These are the state-of-the-art constituency parsing results for the CLE-UTB. 

Table~\ref{tab-13-2} presents the label-wise F-scores for the best performing CNN-BLSTM singe-task constituency parser. The phrase labels with higher frequencies specifically NP, PP, S and VC perform with significantly high F-scores of 89.34, 91.73, 87.82 and 98.00 respectively when evaluated without functional labels. Similarly, these phrase labels have F-scores of 89.40, 91.93, 86.01 and 98.12 with utilizing functional labels. The VC label has the highest F-scored among all others due to an intuitive verbal structure of the language. It also has high appearances in the corpus, therefore, the F-scored for VC makes a significant impact on the overall results. Table~\ref{tab-13-3} shows the label-wise F scores for functional labels. The highest F-score is achieved for the `G' label which is used to mark genitive post-positional phrase. Overall, frequent functional labels perform with high F-scores in single-task parsing experiments.

\begin{figure}[h] 
\centering
\scalebox{0.55}{
\includegraphics{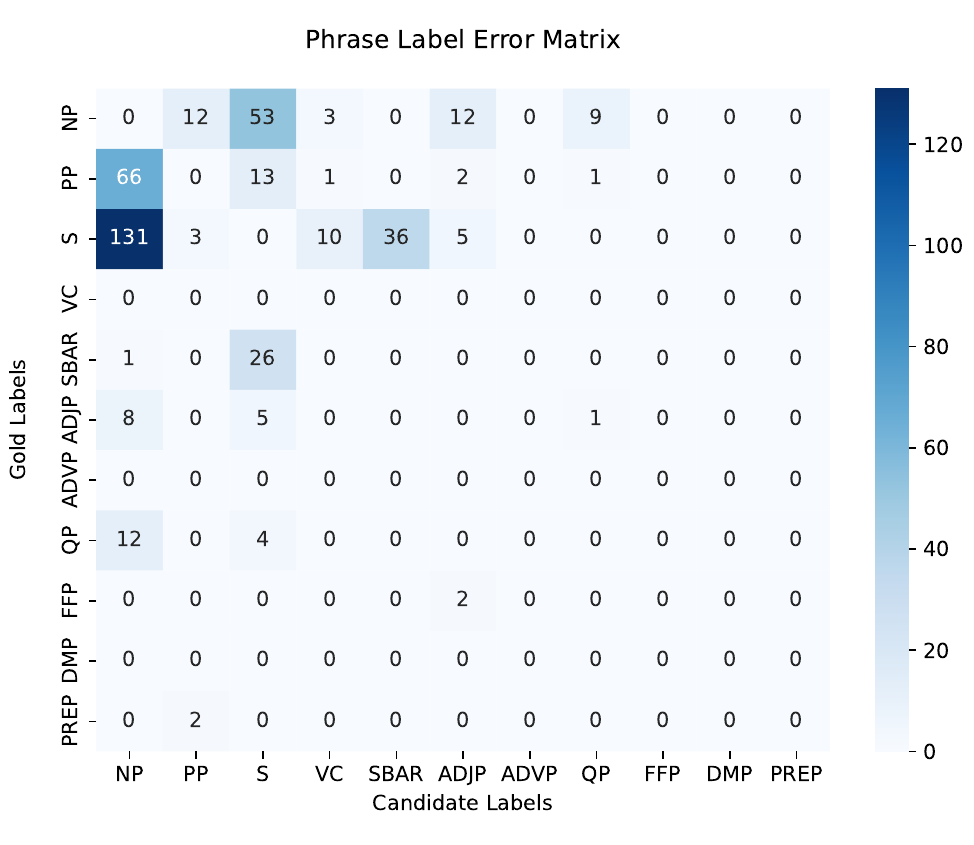}
}
\caption{Phrase label error matrix based on single-task learning paradigm.}
\label{fig-13}
\end{figure}

Figure~\ref{fig-13} shows the error matrices for the best performing single-task learning parser for the phrase labels. It is important to note that the diagram shows only overlapping predicting labels in the gold and candidate labels. The x-axis has labels from candidate set and the gold labels on the y-axis that are present in the evaluation set. The highest confusions are with the `S' and NP followed by the PP and NP labels. PP and NP are closely related in the annotations as each PP constituent contains an NP in it. In the same way, there are confusions between the labels NP and `S' due to their syntactic similarities. 

\begin{figure}[H] 
\centering
\scalebox{0.55}{
\includegraphics{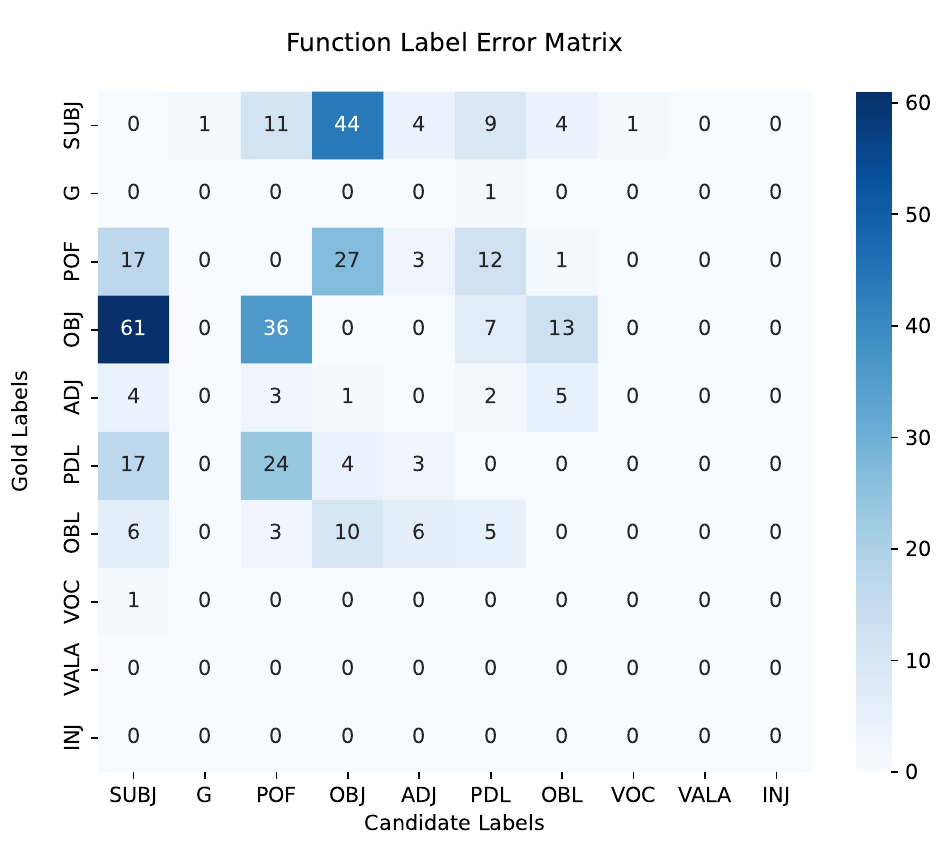}
}
\caption{Function label error matrix based on single-task learning paradigm.}
\label{fig-14}
\end{figure}

Furthermore, Figure~\ref{fig-14} presents confusions among functional labels when predicted by training single-task CNN-BLSTM hybrid model. The highest confusions are between subjects and objects due to the syntactic resemblances. The second largest differences are between objects and POF labels. The Urdu language has a flexible word-order having SOV as the most frequent order with pre-verbal position of objects \cite{ehsan2019analysis}. Similarly, the POF mostly has pre-verbal position that is used to annotate complex predicate structures. This confusion is due to the positional similarity of OBJ and POF. Another prominent disagreement is between POF and PDL. The PDL label is used to annotate copula constructions that usually have pre-verbal positions producing these confusions \cite{ehsan2021development}. However, there are some other minor confusions in the set of predicted candidates. Overall, the constituency parsing results and functional label accuracy are quite promising for the CLE-UTB. 

\begin{figure}[h] 
\centering
\scalebox{0.50}{
\includegraphics{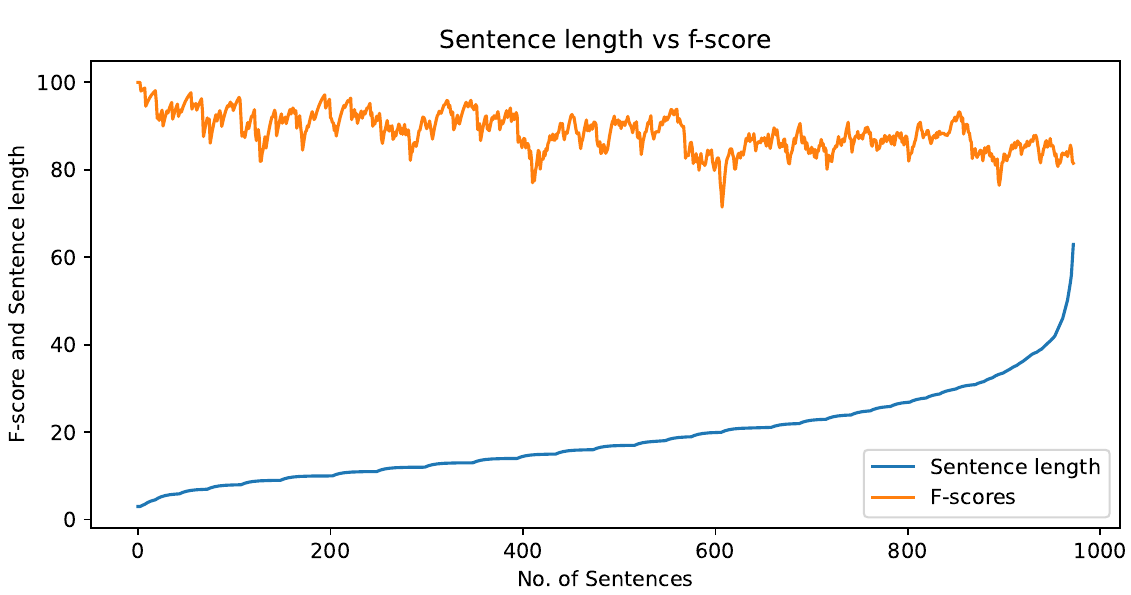}
}
\caption{Constituency parsing results with respect to sentence length.}
\label{fig-15}
\end{figure}

{\color{black}
Figure~\ref{fig-15} presents a graph showing constituency parsing results with respect to sentence lengths. Constituency parsers perform better on shorter sentences, indicating accurate predictions when the syntactic complexity is relatively low, which is consistent with previous studies \cite{ehsan2019analysis}. Sentence lengths are represented with the blue line, and F-scores are represented by an orange color. {\color{black}There is a gradual decline in the F score as the length of the sentence increases}. Despite increased structural complexity and embedding depth, this degradation remains moderate. It is important to note that our parser maintains high performance even for longer sentences, suggesting a strong generalization capability of parsing models and the proposed labeling scheme. This behavior also demonstrates that contextualized representations and training techniques effectively handle the linguistic richness of Urdu, even in structurally dense sentences.
}

\begin{table}[h]
\small
\centering
\caption{Dependency parsing results by using single-task learning paradigm.}
\label{tab-14}
\begin{center}
\setlength\tabcolsep{10pt}
\begin{tabular}{@{}llll@{}}
\hline
Models \& Features & $UAS$ & $LAS$ & $LA$ \\
\thickhline
BLSTM & 79.51 & 74.28 & 87.33 \\
BLSTM + ch-emb & 79.64 & 74.48 & 87.16 \\
BLSTM + ch-emb + pos & 86.75 & 82.55 & 91.45 \\
BLSTM + ch-emb + pos + w2v & 87.59 & 83.59 & 92.05 \\
BLSTM + pos + elmo & 88.75 & 85.13 & 92.84 \\
BLSTM + pos + elmo + w2v & 88.94 & \textbf{85.36} & 93.11 \\
CNN-BLSTM + pos & 86.83 & 82.49 & 91.32 \\
CNN-BLSTM + pos + w2v & 87.41 & 83.42 & 92.01 \\
CNN-BLSTM + pos + elmo & 88.74 & 85.05 & 92.85 \\
CNN-BLSTM + pos + elmo + w2v & 88.87 & \textbf{85.09} & 92.84 \\
\hline 
\end{tabular}
\end{center}
\end{table}

{\color{black}Table~\ref{tab-14} presents the results of the dependency parsing based on single task learning}. As the single-task learning model is similar for both types of syntactic structures, the same features and parameters are used in the training. The labeled attachment score (LAS) is important as it computes the correctness of heads as well as labels. The baseline BLSTM model provides an LAS of 74.28 that is further improved by character embeddings by a point of 0.2. However, POS tags provide a significant improvement of more than eight points with an LAS of 82.55. Word2Vec embeddings further increase LAS scores to 83.59. Using the ELMo embedding, LAS has been enhanced to 85.13 which is quite promising for the converted dependency treebank. Unlike the constituency parser, the dependency parser shows an increase of LAS when trained by combining both word embedding and produces a state-of-the-art LAS of 85.36 on the CLE-UTB. The results outperform the LAS of 81.6 from the famous MatlParser \cite{nivre2007maltparser, ehsan2020dependency} and the LAS of 84.2 from the transition-based BLSTM BIST-Parser \cite{kiperwasser2016simple, ehsan2020dependency}. The CNN-BLSTM hybrid model also produces competitive results; however, for dependency parsing, the BLSTM model performs better.

\begin{table}[h]
\small
\centering
\caption{Label-wise labeled attachment scores (LAS) for dependency parsing using single-task learning paradigm.}
\label{tab-14-2}
\begin{center}
\setlength\tabcolsep{10pt}
\begin{tabular}{@{}lllll@{}}
\hline
Sr.\# & Dependency labels & $Rec.$ & $Pre.$ & $F_1$ \\
\thickhline
1 & acl & 75.26 & 69.96 & 72.51\\
2 & advcl & 64.35 & 78.09 & 70.56\\
3 & advmod & 82.05 & 80.58 & 81.31\\
4 & amod & 88.48 & 88.02 & 88.25\\
5 & aux & 96.39 & 96.73 & 96.56\\
6 & case & 98.66 & 98.79 & 98.72\\
7 & cc & 83.42 & 83.13 & 83.27\\
8 & compound & 84.88 & 83.68 & 84.28\\
9 & conj & 68.40 & 69.5 & 68.95\\
10 & cop & 81.51 & 81.3 & 81.40\\
11 & csubj & 0.00 & 0.00 & nan\\
12 & dep & 88.01 & 91.09 & 89.52\\
13 & det & 92.71 & 90.26 & 91.47\\
14 & discourse & 76.92 & 71.43 & 74.07\\
15 & fixed & 0.00 & 0.00 & nan\\
16 & flat & 70.00 & 70.00 & 70.00\\
17 & iobj & 36.84 & 26.92 & 31.11\\
18 & mark & 91.38 & 90.86 & 91.12\\
19 & nmod & 77.46 & 80.34 & 78.87\\
20 & nsubj & 77.43 & 81.13 & 79.24\\
21 & nummod & 79.87 & 81.70 & 80.77\\
22 & obj & 70.28 & 60.45 & 65.00\\
23 & obl & 83.51 & 83.57 & 83.54\\
24 & punct & 78.27 & 78.17 & 78.22\\
25 & root & 89.21 & 87.06 & 88.12\\
26 & vocative & 62.5 & 66.67 & 64.52\\
27 & xcomp & 41.18 & 50.00 & 45.16\\
\hline 
\end{tabular}
\end{center}
\end{table}

{\color{black}
Table~\ref{tab-14-2} shows the label-wise F-scores for the best performing single-task dependency parser. The scores for the frequent labels are quite promising. {\color{black}However, labels with rare occurrences in the evaluation set such as {\tt csubj} and {\tt fixed} cannot contribute to overall LAS}. The {\tt csubj} has ten samples and the {\tt fixed} does not appear in the evaluation set. It is important to note that supervised methods, specifically techniques based on Deep Neural Networks, are able to perform many NLP tasks but require a substantial amount of training and evaluation data. In our work, both constituency and dependency parsers perform less effective predictions on rarely occurring constructions due to data sparsity. 
}

\begin{table}[h]
\small
\centering
\caption{Confusion matrix of top 10 or more errors for dependency parsing using single-task learning paradigm.}
\label{tab-14-3}
\begin{center}
\setlength\tabcolsep{2pt}
\begin{tabular}{@{}lllllllll@{}}
\hline
$Gold$ & $Cand$ & $Freq.$ & $Gold$ & $Cand$ & $Freq.$ & $Gold$ & $Cand$ & $Freq.$ \\
\thickhline
nsubj & obj & 82 & nmod & obl & 31 & obj & obl & 15\\
nmod & nsubj & 50 & cop & root & 29 & obl & root & 15\\
nmod & obj & 49 & obj & nmod & 29 & nmod & conj & 14\\
nmod & compound & 40 & obl & obj & 23 & root & compound & 14\\
advcl & acl & 37 & obl & nmod & 22 & acl & conj & 13\\
compound & nmod & 37 & compound & nsubj & 19 & advcl & conj & 13\\
compound & obj & 36 & amod & compound & 17 & cop & acl & 12\\
root & cop & 36 & compound & amod & 17 & root & nsubj & 12\\
obj & nsubj & 35 & nummod & det & 17 & nsubj & obl & 11\\
nsubj & compound & 33 & conj & root & 16 & obl & compound & 11\\
nsubj & nmod & 33 & conj & acl & 15 & compound & root & 10\\
obj & compound & 33 & nsubj & root & 15 & iobj & obj & 10\\
\hline 
\end{tabular}
\end{center}
\end{table}

Table~\ref{tab-14-3} further presents the error analysis for dependency labels for the best performing single-task learning model. Due to a large number of labels, the labels with 10 or higher confusions are shown in the table. {\color{black}The highest confusions are among nominal subjects and direct objects because of their syntactic resemblance}. The second and third highest confusions are among the labels {\tt nmod}, {\tt nsubj} and {\tt obj}. The {\tt nmod} label presents nominal modifiers and NPs within post-positional phrases representing genitive cases are mapped on the {\tt nmod} label (Table~\ref{tab-5}). The mapping creates syntactic similarity with nominal subjects and direct objects in the treebank. Similarly, the {\tt nmod} also has significant confusions with the {\tt compound} label. Complex predicate structures are mapped on the {\tt compound} label and the higher portion of the complex predicate structures consists of NPs, hence resulting in these confusions by the parser. There are some minor confusions among the other dependency labels.

\begin{figure}[h] 
\centering
\scalebox{0.55}{
\includegraphics{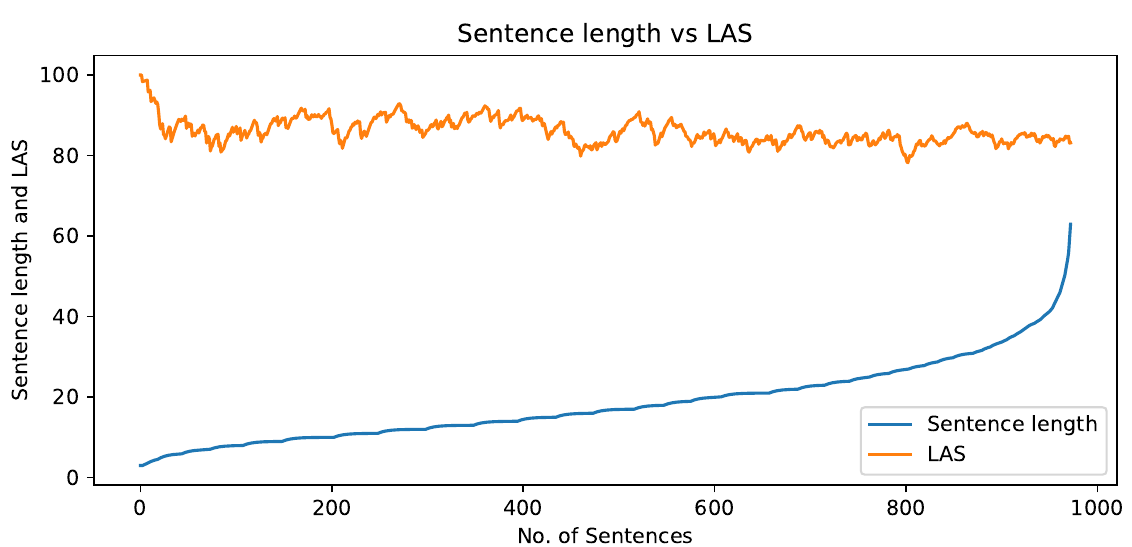}
}
\caption{Dependency parsing results with respect to sentence length.}
\label{fig-17}
\end{figure}

Unlike the constituency parser, the dependency parser performs quite uniformly for sentences of various lengths, as shown in Figure~\ref{fig-17}. The orange line is straighter compared to Figure~\ref{fig-15} depicting the ability of the parser to perform similar for shorter and longer sentences.

The single-task learning-based parsers produce state-of-the-art results for both constituency and dependency structures for the CLE-UTB. However, a multi-task learning-based model is developed to enhance the parsing results by learning cross-structure representations. Table~\ref{tab-15} presents the results of the multi-task learning paradigm trained without utilizing functional labels. Both constituency and dependency parsing results show an upward trend even for the baseline model. The baseline model produces an F-score of 82.49 as compared to 81.41 and an LAS of 74.57 as compared to 74.28 from single-task learning paradigm. Similarly, the feature set further enhances the parsing scores. The multi-task model enhances the F-score for the CLE-UTB to 90.99 when trained with the ELMo Embeddings by training the BLSTM parser. The CNN-BLSTM hybrid model further improves the performance and produces a best F-score of 91.39 for constituency parsing. Without using functional labels, the highest dependency LAS is 85.41 when trained with context-free and contextualized word representations together. It is quite clear that the constituency parsing scores are improved from 90.94 to 91.39 by learning the dependency encodings trained and evaluated without functional labels. There is also a subtle improvement in the dependency results with an LAS of 85.41. However, functional labels could be more useful as they encode the grammatical roles on {\color{black}PS} parse trees.

\begin{table}[h]
\small
\centering
\caption{{\color{black} Multi-task learning based constituency and dependency parsing results without using functional tags.} }
\label{tab-15}
\begin{center}
{\color{black}
\begin{tabular}{@{}lllllll@{}}
\hline
Models \& Features & $Rec.$ & $Pre.$ & $F_1$ & $UAS$ & $LAS$ & $LA$ \\
\thickhline
BLSTM & 81.18 & 83.84 & 82.49 & 79.91 & 74.57 & 87.73 \\
BLSTM+ch-emb & 81.53 & 83.46 & 82.48 & 79.81 & 74.87 & 88.17 \\
BLSTM+ch-emb+pos & 86.49 & 88.43 & 87.45 & 87.01 & 82.79 & 91.5 \\
BLSTM+ch-emb+pos+w2v & 87.03 & 89.08 & 88.04 & 87.52 & 83.42 & 91.81 \\
BLSTM+pos+elmo & 90.38 & 91.62 & 90.99 & 88.3 & 84.45 & 92.51 \\
BLSTM+pos+elmo+w2v & 89.22 & 91.37 & 90.28 & 88.69 & 85.09 & 92.87 \\
CNN-BLSTM+pos & 88.59 & 89.23 & 88.91 & 87.26 & 82.89 & 91.45 \\
CNN-BLSTM+pos+w2v & 88.31 & 89.80 & 89.05 & 87.54 & 83.36 & 91.76 \\
CNN-BLSTM+pos+elmo & 90.67 & 92.12 & \textbf{91.39} & 88.75 & 84.93 & 92.79 \\
CNN-BLSTM+pos+elmo+w2v & 90.17 & 92.13 & 91.14 & 89.13 & \textbf{85.41} & 92.92 \\
\hline
\end{tabular}
}
\end{center}
\end{table}

\begin{table}[h]
\small
\centering
\caption{{\color{black}Multi-task learning based constituency and dependency parsing results by using functional tags.} }
\label{tab-16}
\begin{center}
\setlength\tabcolsep{3pt}
{\color{black}
\begin{tabular}{@{}llllllll@{}}
\hline
Models \& Features & $Rec.$ & $Pre.$ & $F_1$ &F. acc. & $UAS$ & $LAS$ & $LA$ \\
\thickhline
BLSTM & 77.68 & 82.3 & 79.92 & 75.15\% & 79.80 & 74.70 & 87.84\\
BLSTM+ch-emb & 78.97 & 84.17 & 81.49 & 76.72\% & 79.77 & 74.76 & 88.12\\
BLSTM+ch-emb+pos & 82.73 & 87.89 & 85.23 & 80.46\% & 87.14 & 82.93 & 91.62\\
BLSTM+ch-emb+pos+w2v & 83.41 & 88.55 & 85.90 & 81.13\% & 87.42 & 83.38 & 91.93\\
BLSTM+pos+elmo & 86.97 & 91.67 & 89.26 & 84.49\% & 88.75 & 85.23 & 92.98\\
BLSTM+pos+elmo+w2v & 85.47 & 91.18 & 88.23 & 83.46\% & 89.08 & 85.47 & 92.98\\
CNN-BLSTM+pos & 86.63 & 88.63 & 87.61 & 83.43\% & 87.06 & 82.88 & 91.58\\
CNN-BLSTM+pos+w2v & 87.85 & 89.67 & 88.75 & 83.78\% & 87.76 & 83.60 & 91.95\\
CNN-BLSTM+pos+elmo & 90.37 & 91.78 & 91.07 & 86.00\% & 88.68 & 84.90 & 92.73\\
CNN-BLSTM+pos+elmo+w2v & 90.32 & 91.86 & \textbf{91.08} & \textbf{86.23}\% & 89.28 & \textbf{85.69} & 93.20\\

\hline
\end{tabular}
}
\end{center}
\end{table}

\begin{table}[h]
\small
\centering
\caption{Label-wise constituency parsing results from the multi-task learning paradigm.}
\label{tab-17}
\begin{center}
\begin{tabular}{@{}llllllll@{}}
\hline
Sr.\# & Phrase labels & \multicolumn{3}{c}{No functional tags} & \multicolumn{3}{c}{Functional tags}\\ 
& & $Rec.$ & $Pre.$ &$F_1$ & $Rec.$ & $Pre.$ & $F_1$ \\
\thickhline
1 & NP & 89.98 & 90.2 & 90.09 & 89.52 & 90.02 & 89.77\\
2 & PP & 91.14 & 94.04 & 92.57 & 90.56 & 93.23 & 91.87\\
3 & S & 86.13 & 89.10 & 87.59 & 85.58 & 88.10 & 86.82\\
4 & VC & 98.7 & 98.47 & 98.58 & 98.47 & 97.92 & 98.19\\
5 & SBAR & 83.36 & 86.02 & 84.67 & 83.02 & 84.76 & 83.88\\
6 & ADJP & 78.26 & 82.26 & 80.21 & 77.49 & 82.79 & 80.05\\
7 & ADVP & 91.14 & 89.26 & 90.19 & 91.14 & 88.16 & 89.63\\
8 & QP & 70.75 & 76.47 & 73.5 & 62.59 & 83.64 & 71.60\\
9 & FFP & 61.9 & 92.86 & 74.29 & 71.43 & 100 & 83.33\\
10 & DMP & 20.0 & 20.0 & 20.0 & 40.0 & 40.0 & 40.0\\
11 & PREP & 0.00 & nan & nan & 0.00 & nan & nan\\
\hline
\end{tabular}
\end{center}
\end{table}

Table~\ref{tab-16} presents the multi-task learning-based parsing scores by utilizing functional labels in experiments. The CNN-BLSTM hybrid model produces a best LAS of 85.47 when trained with Word2Vec and ELMo embeddings. The constituency results are not much improved when trained with functional labels as compared to single-task learning model. The functional labels encode the dependency information in the {\color{black}PS} trees therefore, the presence of functional labels in the constituency structure parse trees does not improve leaning capability of the model. On the other hand, functional labels contribute in the generation of larger set of sequential labels. However, the accuracy of the functional labels improved significantly incorporating {\color{black}DS}-based encodings in the model. The accuracy for functional labels is enhanced by 1.51 points from 84.72\% to 86.23\%. Overall, the constituency parsing is improved by using {\color{black}DS} encodings trained without using functional labels and the dependency parsing results are enhanced by incorporating constituency structure encodings along with functional labels.

\begin{table}[h]
\small
\centering
\caption{Label-wise LAS for dependency parsing from the multi-task learning paradigm.}
\label{tab-18}
\begin{center}
\begin{tabular}{@{}llllllll@{}}
\hline
Dep. labels & $Rec.$ & $Pre.$ & $F_1$ & Dep. labels & $Rec.$ & $Pre.$ & $F_1$ \\
\thickhline
acl & 74.23 & 70.35 & 72.24 & fixed & 0.00 & nan & nan\\
advcl & 64.81 & 78.21 & 70.88 & flat & 70.00 & 58.33 & 63.63\\
advmod & 81.59 & 82.91 & 82.24 & iobj & 31.58 & 50.00 & 38.71\\
amod & 87.85 & 89.64 & 88.74 & mark & 90.23 & 89.54 & 89.88\\
aux & 96.53 & 96.6 & 96.56 & nmod & 77.99 & 81.09 & 79.51\\
case & 98.28 & 98.15 & 98.21 & nsubj & 82.61 & 79.01 & 80.77\\
cc & 83.60 & 83.9 & 83.75 & nummod & 82.11 & 87.12 & 84.54\\
compound & 86.65 & 82.61 & 84.58 & obj & 64.26 & 67.23 & 65.71\\
conj & 72.99 & 71.63 & 72.30 & obl & 83.51 & 84.29 & 83.90\\
cop & 80.73 & 81.58 & 81.15 & punct & 79.81 & 79.55 & 79.68\\
csubj & 0.00 & nan & nan & root & 89.31 & 87.69 & 88.49\\
dep & 88.39 & 91.83 & 90.08 & vocative & 68.75 & 75.86 & 72.13\\
det & 94.29 & 91.03 & 92.63 & xcomp & 52.94 & 52.94 & 52.94\\
discourse & 69.23 & 47.37 & 56.25 & --- & --- & --- & --- \\
\hline 
\end{tabular}
\end{center}
\end{table}

Table~\ref{tab-17} and Table~\ref{tab-18} show label-wise constituency and dependency parsing results achieved from the best performing multi-task learning models. The parsing scores for the core and frequent labels have been enhanced by utilizing cross-structure encodings. For the constituency structure, improvements are observed for NP, PP, ADJP and DMP labels with increase of 0.75, 0.84, 2.32 and 40.0 points respectively. Similarly, for the {\color{black}DS}, prominent improvements are analyzed for some of the core labels {\tt nsubj}, {\tt obj}, {\tt nmod} and {\tt iobj} with increase of 1.53, 0.71, 0.64 and 7.60 points respectively.

\begin{figure}[h]
\centering
\scalebox{0.55}{
\includegraphics{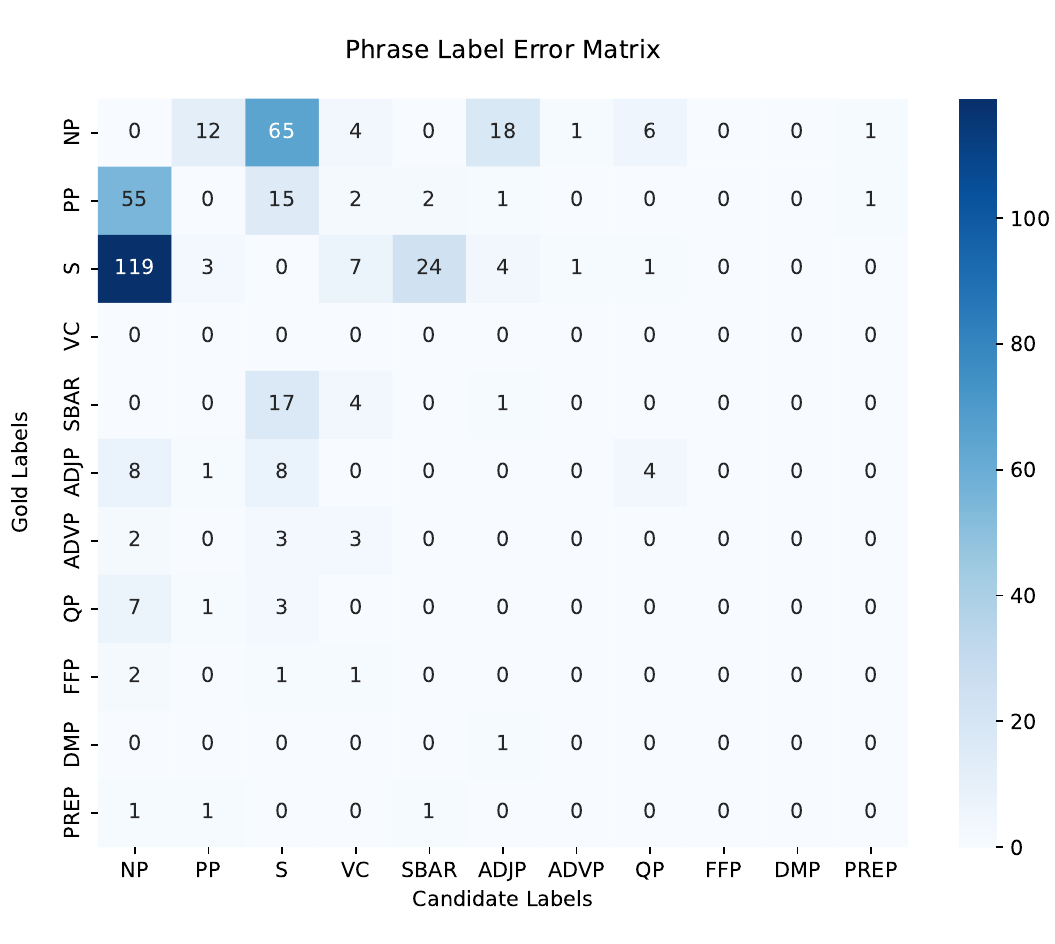}
}
\caption{Phrase label error matrix based on the multi-task learning paradigm.}
\label{fig-18}
\end{figure}

\begin{figure}[H] 
\centering
\scalebox{0.60}{
\includegraphics{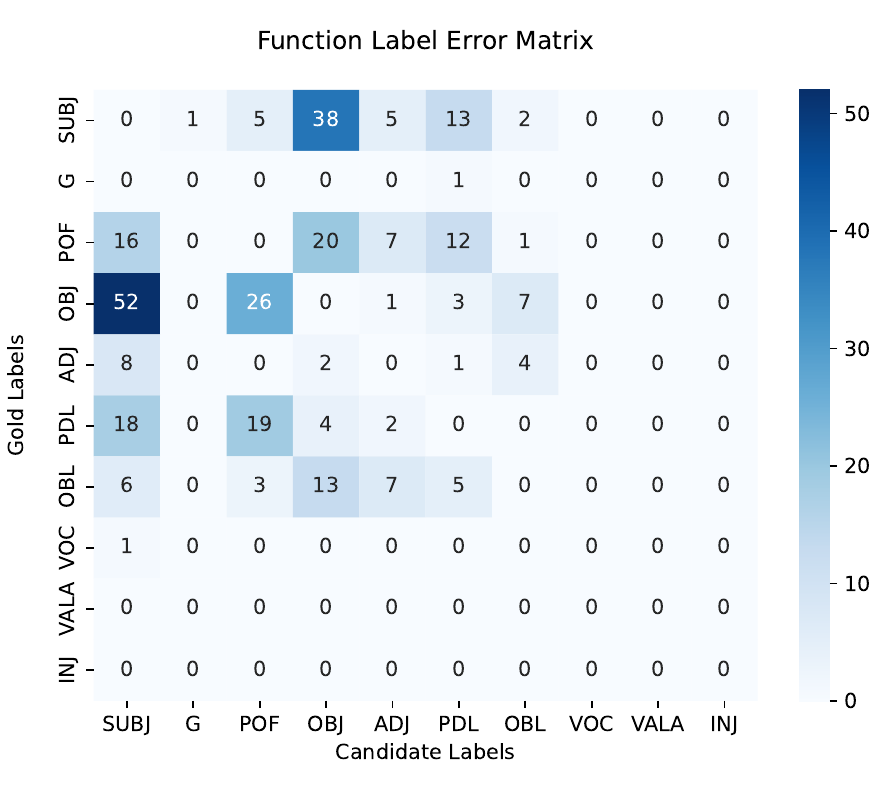}
}
\caption{Function label error matrix based on the multi-task learning paradigm.}
\label{fig-19}
\end{figure}

Figure~\ref{fig-18} and Figure~\ref{fig-19} demonstrate the enhancement of the parsing results by showing reduced confusions between phrase and functional labels. The errors between `S' and NP labels are reduced from 131 to 119. Similarly, the confusions between PP and NP labels are decreased from 66 to 55. However, an increase of errors is observed between NP and `S' labels, overall, the multi-task learning paradigm is helpful to achieve improved results with minimized confusions between labels. Similarly, the difference of functional labels in the evaluation set are reduced. The highest disagreement was observed between OBJ and SUBJ with a value of 61 that is reduced to 52 and the confusions between SUBJ and OBJ are minimized from 44 to 38. The label-wise errors are reduced for all the functional labels by employing multi-task learning models with a reduction of 12.4\%.

Table~\ref{tab-19} shows confusions for the dependency parser for ten or higher disagreements. The highest confusions of 82 were observed with the {\tt nsubj} and {\tt obj} label that are reduced to 37 in the multi-task learning paradigm. Some of the confusion values show the differences in the opposite direction. However, total number of disagreements in the top 36 labels are reduced from 901 to 861 with an error reductions of 4.4\%.

\begin{table}[h]
\small
\centering
\caption{Confusion matrix of top 10 or more errors for dependency parsing by Multi-task learning paradigm.}
\label{tab-19}
\begin{center}
\setlength\tabcolsep{3pt}
\begin{tabular}{@{}lllllllll@{}}
\hline
$Gold$ & $Cand$ & $Freq.$ & $Gold$ & $Cand$ & $Freq.$ & $Gold$ & $Cand$ & $Freq.$ \\
\thickhline
obj & nsubj & 59 & nsubj & nmod & 27 & compound & root & 14\\
nmod & nsubj & 57 & cop & root & 26 & conj & root & 14\\
nmod & compound & 44 & compound & obj & 25 & cop & acl & 14\\
nsubj & obj & 37 & obl & nmod & 25 & advcl & conj & 14\\
obj & compound & 37 & compound & nsubj & 24 & obl & root & 13\\
advcl & acl & 37 & amod & compound & 23 & obl & nsubj & 13\\
nmod & obl & 37 & obl & obj & 22 & acl & conj & 12\\
root & cop & 35 & nummod & det & 18 & obj & obl & 11\\
obj & nmod & 35 & root & nsubj & 17 & nsubj & obl & 10\\
nmod & obj & 32 & root & compound & 16 & acl & advcl & 10\\
compound & nmod & 28 & nsubj & root & 15 & nmod & conj & 10\\
nsubj & compound & 27 & obl & compound & 14 & acl & cop & 9\\
\hline 
\end{tabular}
\end{center}
\end{table}

We conducted multiple parsing experiments by employing single-task and multi-task learning paradigms. Single-task learning models are trained for the constituency and dependency parsing independently. Various features are incorporated with the single-task learning models and state-of-the-art parsing results are achieved for the CLE-UTB. Furthermore, multi-task learning models are trained to incorporate cross-structure representations in the models. The cross-structure encodings were quite helpful to enhance parsing results. The experiments are conducted by including the functional labels as well as without them. The constituency parsing results are enhanced by utilizing the dependency encodings in the models. However, the dependency parsing scores are enhanced by incorporating the functional labels in the constituency treebank. The accuracy of the functional labels is also improved by the multi-task learning models. Overall, the parsing results are evident that learning cross-structure representations are quite helpful for parsing task. The contextualized word representations are also helpful in producing improved results by performing transfer learning for small to medium-sized annotated data sets.
We conducted multiple parsing experiments by employing single-task and multi-task learning paradigms. Single-task learning models are trained for the constituency and dependency parsing independently. Various features are incorporated with the single-task learning models and state-of-the-art parsing results are achieved for the CLE-UTB. Furthermore, multi-task learning models are trained to incorporate cross-structure representations in the models. The cross-structure encodings were quite helpful to enhance parsing results. The experiments are conducted by including the functional labels as well as without them. The constituency parsing results are enhanced by utilizing the dependency encodings in the models. However, the dependency parsing scores are enhanced by incorporating the functional labels in the constituency treebank. The accuracy of the functional labels is also improved by the multi-task learning models. Overall, the parsing results are evident that learning cross-structure representations are quite helpful for parsing task. The contextualized word representations are also helpful in producing improved results by performing transfer learning for small to medium-sized annotated data sets.


\section{Conclusion}
\label{sec-7}

This work presents two syntactic parsing paradigms to parse a {\color{black}morphologically rich} language Urdu. First, the single-task learning paradigm along with several features including character embeddings, POS tags, context-free and contextualized word representations. Second, the multi-task learning paradigm that performs the parsing task by learning cross-structure representations. A {\color{black}PS} treebank (CLE-UTB) is transformed to the {\color{black}DS}. A language-dependent head-word model and a constituency to dependency label mapping scheme are devised followed by several post-conversion rules. The single-task learning paradigm produced state-of-the-art constituency and dependency parsing results for the CLE-UTB when trained with contextualized word representations. The character-based contextualized representations are quite capable of learning the morphology and syntactic information of the language. The multi-task learning paradigm further enhanced the parsing results by learning cross-structure representations. The results are evident that cross-structure encodings are useful to improve the syntactic parsing.

\section*{Acknowledgments}
The authors extend their appreciation to Umm Al-Qura University, Saudi Arabia, for funding this research.

\nolinenumbers

%
%
%
\bibliography{bibliography}





\end{document}